\documentclass[jorn]{IEEEtran}

\usepackage[american]{babel}

\usepackage[caption=false]{subfig}
\usepackage{graphicx}
\usepackage{amsmath,amssymb}
\usepackage{algorithm,algorithmic}
\usepackage{multirow}
\usepackage{diagbox}
\usepackage{mathtools}
\usepackage{url}
\usepackage{xcolor}
\usepackage{colortbl}
\usepackage{changepage}

\usepackage[pagebackref=false,breaklinks=true,colorlinks,bookmarks=false]{hyperref}

\newcommand{\mbf}[1]{\mathbf{#1}}

\newcommand{\red}[1]{\textcolor{black}{#1}}
\newcommand{\blue}[1]{\textcolor{black}{#1}}
\newcommand{\redt}[1]{\textcolor{black}{#1}}
\newcommand{\bluet}[1]{\textcolor{black}{#1}}
\newcommand{\pink}[1]{\textcolor{black}{#1}}
\newcommand{\gassian}{\mathcal{N}}

\definecolor{mygray}{gray}{0.85}

\DeclareGraphicsExtensions{.pdf,.jpeg,.png,.eps}

\begin{document}

\title{Group Re-Identification with Multi-grained Matching and Integration}

\author{Weiyao Lin,~Yuxi Li,~Hao Xiao,~John See,~Junni Zou,~Hongkai Xiong,~Jingdong Wang and Tao Mei%





\thanks{
This work was supported in part by the National Natural Science Foundation of China under Grants61529101, 61425011, 61720106001, in part by Shanghai 'The Belt and Road' Young Scholar Exchange Grant (17510740100), and in part by CREST Malaysia (No. T03C1-17).}
\thanks{W. Lin, Y. Li, H. Xiao and H. Xiong are with the Department of Electronic Engineering, Shanghai Jiao Tong University, China (email: \{wylin, lyxok1, alexinsjtu, xionghongkai\}@sjtu.edu.cn). }
\thanks{J. See is with the Faculty of Computing and Informatics, Multimedia University, Malaysia (email: johnsee@mmu.edu.my).}
\thanks{J. Zou is with the Department of Computer Science, Shanghai Jiao Tong University, China (email: zoujn@cs.sjtu.edu.cn).}
\thanks{J. Wang is with Microsoft Research Asia, Beijing, China (email: jingdw@microsoft.com).}
\thanks{T. Mei is with JD AI Research, Beijing, China (email: tmei@jd.com).}
}

\maketitle

\begin{abstract}
\blue{The task of} re-identifying groups of people \red{under} different camera views is an important yet less-studied problem. Group re-identification (Re-ID) is a 
very challenging \blue{task} since it is not only \blue{adversely affected} 
by \red{common issues in traditional single object Re-ID problems such as} viewpoint and human pose variations, but it also suffers from \blue{changes} in group layout and group membership. 
\blue{In this paper,}
we propose \blue{a novel concept of group granularity by characterizing a group image by} 
multi-grained objects:
individual persons and sub-groups of two and three people \blue{within} a group.
\blue{To achieve robust group Re-ID,}
\red{we \blue{first introduce multi-grained representations which can be extracted via} the development of two separate schemes, i.e. one with hand-crafted descriptors and \blue{another} with deep neural networks.} 
\blue{The proposed representation seeks to characterize both appearance and spatial relations of multi-grained objects, and is further equipped with importance weights which capture variations in intra-group dynamics.}
Optimal group-wise matching \blue{is facilitated}
by a multi-order matching process
\blue{which in turn, }
dynamically updates the importance weights 
\blue{in iterative fashion.}
\blue{We evaluated on three multi-camera group datasets containing complex scenarios and large dynamics, with experimental results demonstrating the effectiveness of our approach.} 
\end{abstract}

\begin{IEEEkeywords}
Re-identification, Group Re-ID, Multi-grained representation, Group-wise matching
\end{IEEEkeywords}

\section{Introduction\label{section:introduction}}

Person re-identification (Re-ID) aims at matching and identifying pedestrians across non-overlapping camera views. This task is increasingly important in visual surveillance and has attracted much attention in recent research \cite{Ding2015deep,chen2016deep,shenyang}. However, most research focused on individual person Re-ID, while the Re-ID of groups of people are seldom studied. In practice, since most events (e.g., moving, fighting or violent actions) could be performed \red{within distinct groups} instead of between individuals, 
\bluet{it is essential to
identify} groups rather than \red{single persons} when analyzing events across cameras~\cite{zheng,cai}. Therefore, it is non-trivial to obtain reliable group matching across \redt{different camera views}.

\begin{figure}[tb!]
	\centering
	\captionsetup[subfigure]{labelformat=empty}
	\begin{minipage}{0.14\textwidth}
		\subfloat[(a)] {\includegraphics[width=1.05\linewidth,height=0.83\linewidth]{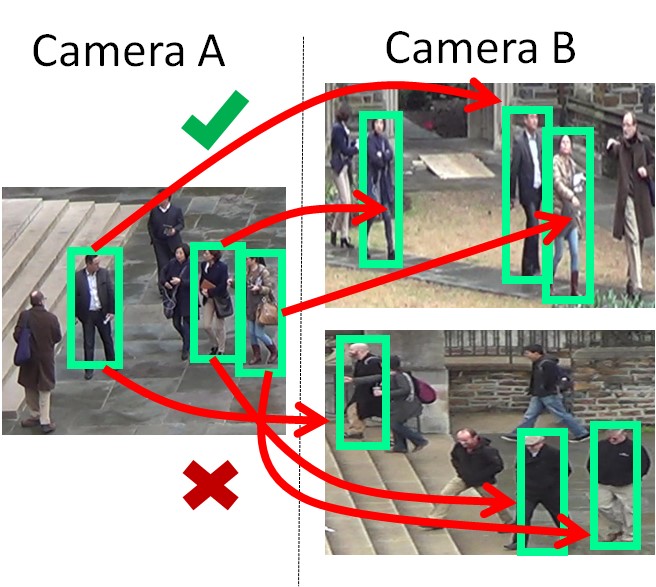}
			\label{fig1a}}  \\
		\vspace{-6mm}
		\subfloat[(b)] {\includegraphics[width=1.05\linewidth,height=0.83\linewidth]{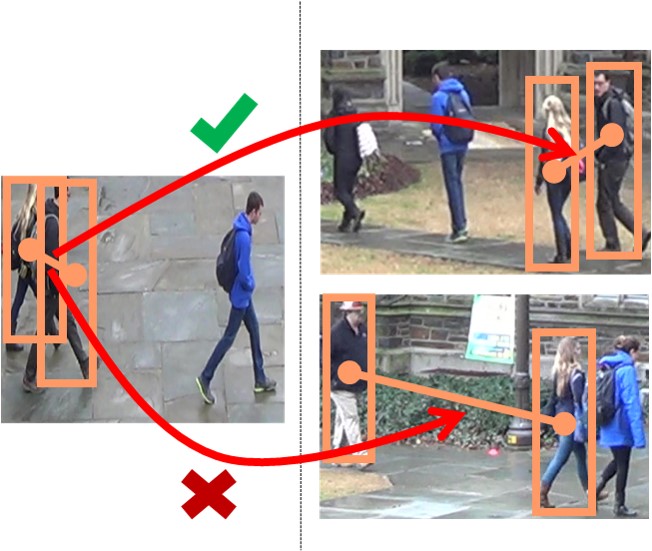}
			\label{fig1b}}
	\end{minipage}
	\hspace{2.1mm}
	{\vrule width0.6pt}
	\hspace{0.1mm}
	\begin{minipage}{0.30\textwidth}
		\subfloat[(c)]{\includegraphics[width=1.0\linewidth,height=0.8\linewidth]{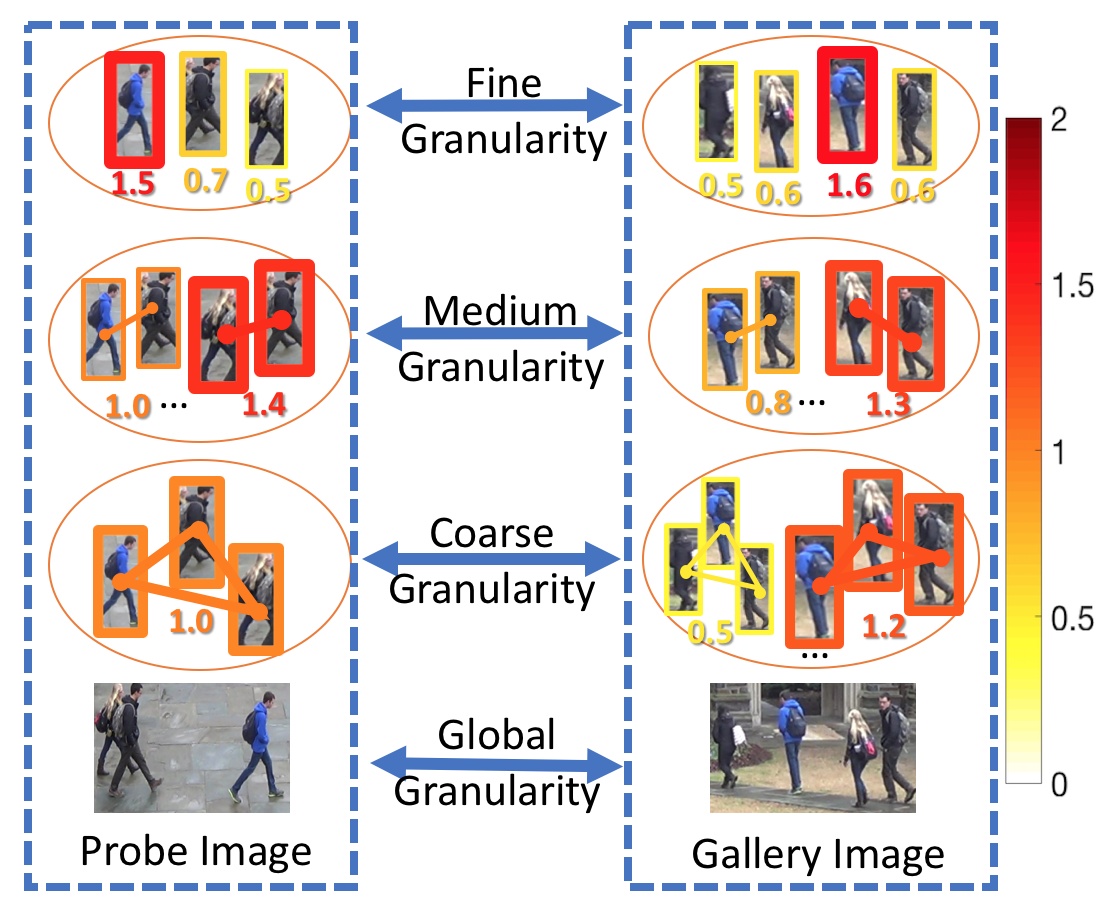}
			\label{fig1c}}
	\end{minipage}
	\caption{(a)-(b) Left: Probe groups in camera $A$; Right: The correctly matched groups (top) and incorrectly matched groups (bottom) in gallery camera $B$. (c): Illustration of multi-grained information for group Re-ID. The colored lines and rectangles in (c) indicate the importance weights for individuals and people subgroups. (Best viewed in color)
	}\label{fig1}
	\vspace{-2mm}
\end{figure}

\bluet{In group Re-ID,} there are two more basic challenges besides viewpoint changes and human pose variations~\cite{zhaorui2,shenyang}, \bluet{both inherent issues} for individual person \bluet{case}. \bluet{These challenges are as follows:}
(i) Group layout change:
The layout of people in a group \bluet{are largely unconstrained} 
across different camera views. Due to the dynamic movements of people, the relative positions of people in a group may have large differences in two camera views (cf. Fig.~\ref{fig1a}).
(ii) Group membership change: people may often join or leave a group (cf. Fig.~\ref{fig1b}).

\begin{figure*}[tb!]
	\centering
	\vspace{-2mm}
	\includegraphics[height=0.28\textwidth,width=\textwidth]{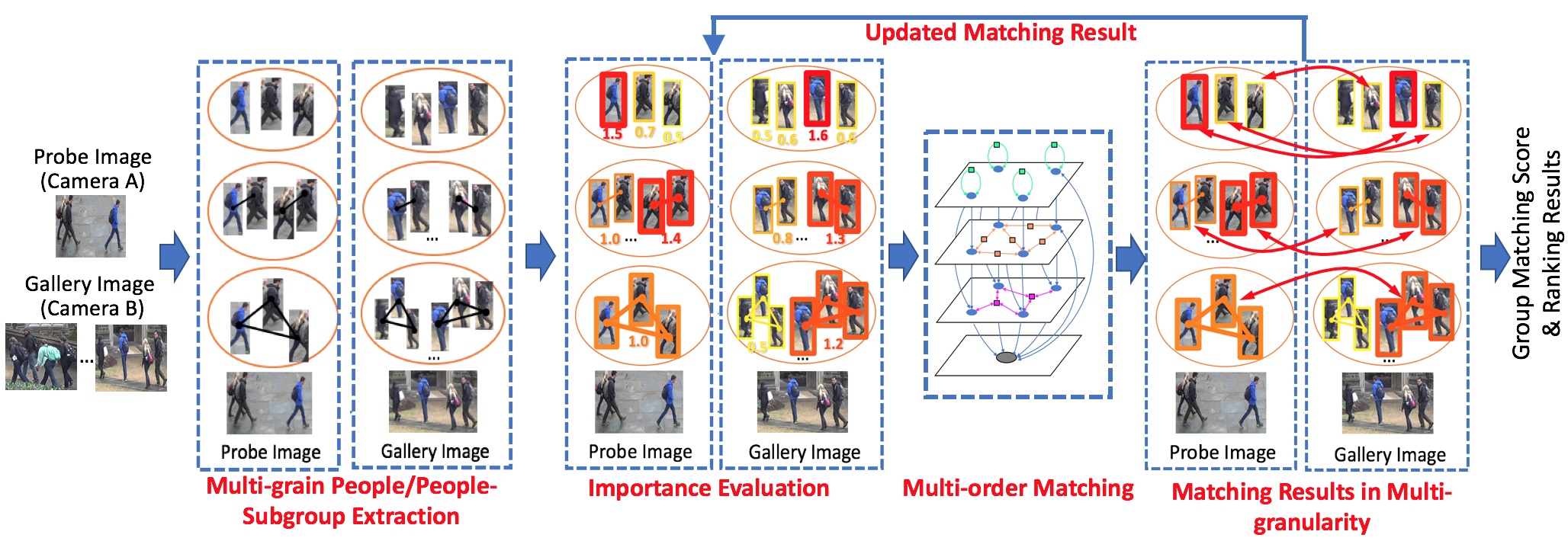} 
	\vspace{-2mm}
	\caption{Framework of the proposed group Re-ID approach, which is composed of three sequential parts, i.e. multi-grained feature extraction, importance evaluation and multiple order matching. (Best viewed in color)}\label{fig:framework}
	\vspace{-2mm}
\end{figure*}

Most existing methods, e.g.,~\cite{zheng,cai,lisanti2017group},
view the input group image as an entire unit
and extract global/semi-global features
without explicitly doing individual people matching
and considering layout changes
to perform group-wise matching~\cite{zheng,cai}.
A recent study~\cite{icip16} attempts to use descriptors of local patches
to partially handle layout change and membership change.

In this paper, we introduce the idea of \emph{group granularity}
and characterize a group image
by \blue{\emph{multi-grained objects}}. \redt{\bluet{By defining crowds} with large membership overlap as of the same group, a group can be depicted with multi-grained objects: 
}
fine-grained objects \blue{are} formed by a single person,
medium-grained objects \blue{are} formed by a group of two people,
coarse-grained objects \blue{are} formed by a group of three people,
and global-grained object \blue{consists of all persons in the group.}
We argue that \blue{characterization of objects from multiple granularities} is helpful to \redt{enhance the invariance of descriptors \bluet{towards} changes in \bluet{group} membership and layout}.

\blue{We refer to}
the example in Fig.~\ref{fig1a} \blue{for better clarity on the need for group granularity.}
Due to the large layout variation and camera viewpoint changes, the same group shows large global appearance differences in two camera views.
The Re-ID performance is poor if global features are merely adopted for the entire group for re-identification (cf. bottom-right image in Fig.~\ref{fig1a}). This issue can be resolved
if we include information of finer group granularity (e.g., individual persons).
On the other hand, merely using the information of individual people is also not always reliable. An example is shown in Fig.~\ref{fig1b} where two groups in Camera $B$ which contain visually similar group members
to the probe group in Camera $A$.
In this case, the information at medium-level granularity (e.g., subgroups of two people) would be useful.

Our approach leverages on the representations of \blue{multiple granularities}, also termed as multi-grained objects (cf. Fig.~\ref{fig1c}),
for group re-identification.
In addition, motivated by the observations that
groups in different cameras may be interfered by group member variation, occlusion, and mismatching,
and that multi-grained objects have different reliabilities on Re-ID performances,
we propose to introduce the dynamic updated importance weights \red{to explicitly model the different characterization power of each objects in different granularity and further}
improve the group re-identification performance.

\red{Meanwhile, due to the strong ablility of convolutional neural networks to extract local invariant features, some deep learning based methods have achieved unprecedented performance in vision recognition tasks in recent years \cite{He2015Deep,Simonyan2014Very}. Inspired by these works, many deep learning techniques have been applied to the person Re-ID task \cite{Chen2017Beyond,Hermans2017Trihard} and are demonstrated to
outperform some traditional pipelines, which are \blue{typically reliant on} manually designed feature representations and metric learning algorithms. Nevertheless, few works utilize deep learning methods for group re-identification.
Considering the 
large variation in illumination, \redt{membership change in crowds} and pose transformation in group Re-ID dataset, \blue{there is sufficient motivation to study} 
the performance of deep CNNs on the group Re-ID task. 
}

\red{\blue{For comprehensiveness}, we 
introduce two independent pipelines for feature extraction. One is a combination of traditional handcrafted algorithms while the other is based on a multi-task convolutional neural network. \blue{Both pipelines} 
extract \blue{their own set of} appearance and spatial relation features of different granularities.
Our experiments \blue{convincingly} show that our group Re-ID framework is able to achieve state-of-the-art results on different datasets with either hand-crafted features or deep convolutional features.}

In summary,
our contributions \blue{are four-fold}:
\begin{enumerate}
	\item We introduce multi-grained representations for group images
	to better handle \blue{changes} in group layout and membership,
	\blue{coupled with} a dynamic weighting scheme for better person matching.
	\vspace{1mm}
	\item We solve the group-wise matching problem by using a multi-order matching algorithm that integrates multi-grained representations and combines the information of both matched and unmatched objects to achieve a more reliable matching result.
	\vspace{1mm}
	\red{\item We propose two schemes to 
	extract \blue{appearance and spatial relation features} for the multi-grained representation: one based on typical hand-crafted features, the other on \blue{deep} CNN features. \blue{For the latter, a new multi-task integrated CNN is designed for this specific purpose.}}
	\vspace{1mm}
	\item We create two challenging group Re-ID datasets with large group membership and layout variations. The existing group Re-ID datasets \blue{consists of }
	relatively small \blue{group sizes and group layout, which are less realistic in real-world scenarios.} 
\end{enumerate}

\section{Related Works and Overview}
\label{section:related_work}
Person Re-ID has been studied for years. Most of them focus on developing reliable features \cite{zhang2017video,ma2014covariance}, deriving accurate feature-wise distance metric \cite{mirror,zhang2016learning}, and handling local spatial misalignment between people \cite{shenyang,tan2016dense}. Some recent research works extend Re-ID algorithms to more object types (e.g., cars \cite{shen2017learning}) or more complex scenarios (e.g., larger camera numbers \cite{multi-camera}, long-term videos \cite{zhengliang1}, untrimmed images \cite{zheng2016person,xiao2017ian}).

 \red{During recent years, some deep learning based methods have emerged to solve the problem of single person Re-ID. These methods have a strong ability to extract rich invariant features from images. \blue{A number of works} \cite{Li2014DeepReID,Chen2016A} exploit CNNs for person Re-ID by \blue{exploiting pairwise labels from positive and negative sample pairs in a variety of network architectures.} 
 \blue{More recent works} \cite{Li2017Person,Chen2017Beyond,Zhang2017AlignedReID} \blue{are inclined to} equip CNNs with triplet loss~\cite{Schroff2015FaceNet}, which has shown to perform exceedingly well \cite{Hermans2017Trihard,liao2017triplet} by learning a representative feature embedding space \blue{which facilitates} a distance metric.}

However, most existing works focus on the Re-ID of individual person; as such, the group-level Re-ID problem is seldom considered. Since group Re-ID contains significant group layout changes and group membership variations, it introduces new challenges and \blue{a proliferation of} 
information \blue{that requires encoding as }
compared to 
scenarios addressed by single person Re-ID methods. Although some works \cite{group-info1,group-info2} introduce people or group interaction into Re-ID process, they are only targeted at improving the Re-ID performance of a single person. The characteristics of groups are still less considered and not fully modeled.

Only a few works \blue{have been proposed} to address group Re-ID tasks \cite{zheng,cai,icip16,lisanti2017group}. Most of them develop global or semi-global features to perform group-wise matching. For example, Cai et al.~\cite{cai} proposed a discriminative covariance descriptor to capture the global appearance \& statistic properties of group images. Zheng et al.~\cite{zheng} segmented a group image into multiple ring regions and derived semi-global descriptors for each region. Lisanti et al.~\cite{lisanti2017group} combined sparsity-driven descriptions of all patches into a global group representation. Since global or semi-global features \blue{are unlikely to} 
capture information of local interaction in groups, they \blue{may} have limitations in handling complex scenarios with significant \blue{group appearance variations caused by pose and background interference.}

Recently, Zhu et al. \cite{icip16} developed a local-based method which performs group Re-ID by selecting proper patch-pairs and conducting patch matching between cross-view group images. However, in order to reduce patch mismatches, this method includes prior restrictions on vertical misalignments. This limits their capability in handling significant group layout changes or group member variations.

Our approach differs from the existing group Re-ID works in two aspects: (1) The existing works perform Re-ID with 
\blue{information derived from single granularity} (i.e., either global or patch level information). Comparatively, our approach leverages multi-grained information to fully capture the characteristics of a group. (2) Our approach does not include any prior restrictions on spatial misalignments, which is able to handle arbitrary group layout changes or group member variations.

\vspace{.1cm}
{\bf Overview of our approach.} Given the probe group image captured from one camera, 
our goal is to find the matched group images
from a set of gallery group images captured from another camera.
We represent each group image
by a set of multi-grained objects,
and then \blue{proceed to} extract features 
\red{by a combination of hand-crafted descriptors, or by a forward pass on a multi-task CNN
}.
\blue{With these features,} the matching process computes the static and dynamic importance weights of multi-grained objects \blue{between} 
the probe and gallery images. 
Then, a multi-order matching algorithm
computes intermediate matching results, which are used to update the dynamic importance weights.
We perform these two stages in iterative fashion, with final matching results \blue{obtained at convergence.}
The entire framework is shown in Fig.~\ref{fig:framework}.

\section{Multi-grained Representation\label{section:feature extract}}
A group image $I$
contains a set of people:
$\mathcal{B} = \{b_1, b_2, \dots, b_N\}$,
where $N$ is the number of people
and $b_i$ (or simply denoted by $i$
for presentation clarity)
corresponds to the person bounding box.
The representation is computed
by building multi-grained objects (people/subgroups):
1) Fine granularity, including objects of an individual person,
$\mathcal{O}_1 = \{i|i=1, \dots, N\}$;
2) Medium granularity, including objects of two-people subgroups,
$\mathcal{O}_2 = \{(i_1, i_2)| i_1, i_2=1, \dots, N, i_1 \neq i_2\}$;
3) Coarse granularity, including objects of three-people subgroups,
$\mathcal{O}_3 = \{(i_1, i_2, i_3)| i_1, i_2, i_3=1, \dots, N, i_1 \neq i_2 \neq i_3\}$;
and 4) Global granularity, referring to the entire group,
$\mathcal{O}_g = \{(1, 2, \dots, N)\}$.
In the 
\blue{cases where} there are only two people in the group image,
we simply let $\mathcal{O}_2$ be the coarse granularity. 

\textbf{\bluet{Choice of granularities.}} \redt{We adopt four \bluet{levels} 
of granularity 
because the combination of \bluet{three distinct levels and a global level} is sufficient to characterize both the global appearance and local layout of crowds, \bluet{besides for tractability reasons.}}
The fine granularity helps reduce the confusion in the global appearance
\blue{when encountering} large layout or group member changes,
while the medium and coarse granularities 
can help resolve ambiguous individual person matches in the fine granularity by incorporating local layout or co-occurrence information in a group.

\begin{figure*}[ht]
	\centering
	\hspace{2mm}
	\subfloat[]{\includegraphics[width=0.58\textwidth]{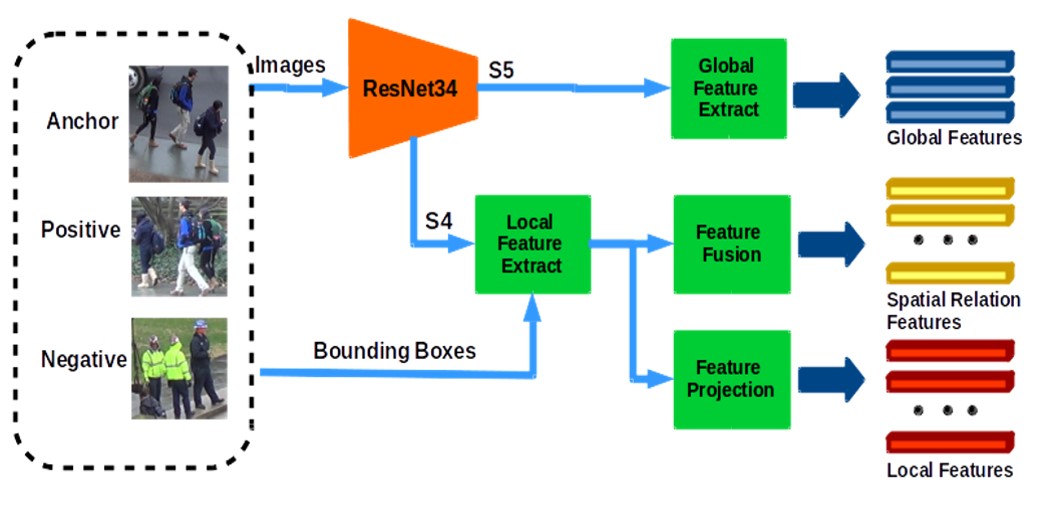}\label{fig:cnn}}
	\hspace{2mm}
	{\vrule width0.6pt}
	\hspace{2mm}
	\subfloat[]{\includegraphics[width=0.28\textwidth]{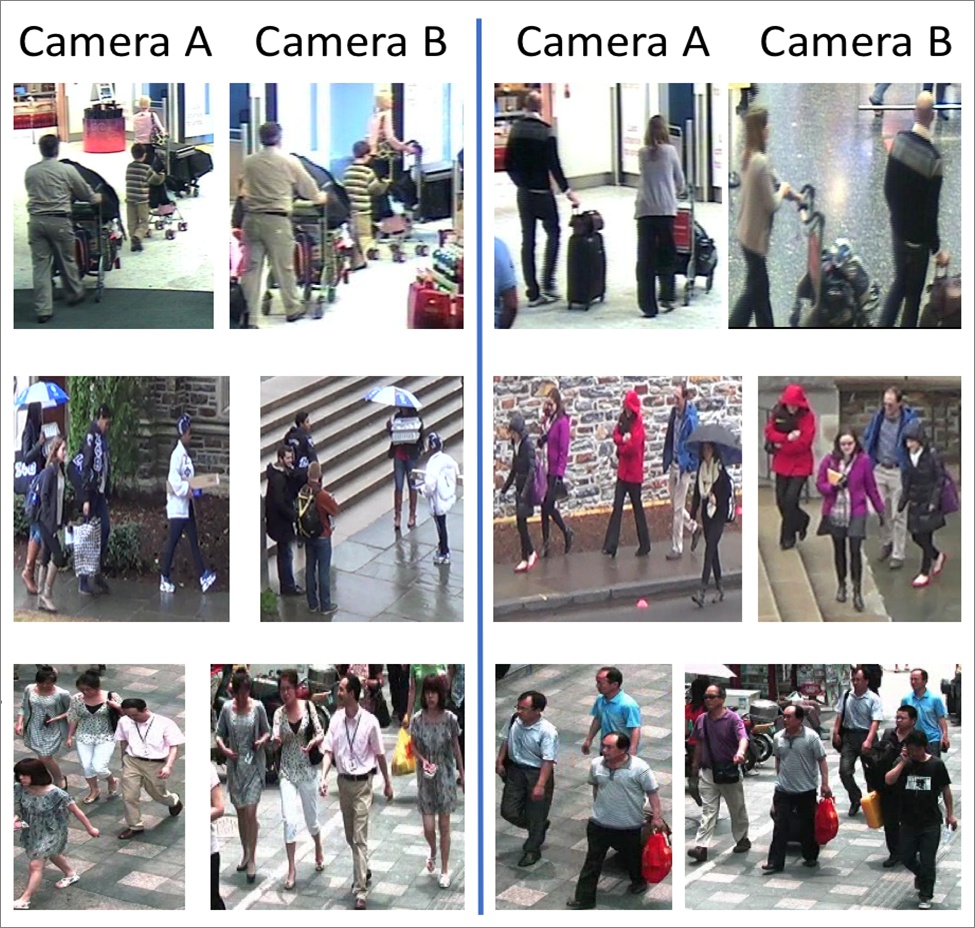}
		\label{fig:datasets}}
	\caption{(a) Overview of our multi-task CNN for feature extraction from input group images.(best viewed in color), The dotted rectangle denotes a triplet of input images, consisting of the Anchor, Positive and Negative samples. S4 and S5 denote the 4-th and 5-th stage of ResNet-34 CNN. (b) Examples from datasets used in our experiments. The first row is from \emph{i-LID MCTS}; The second and third rows are from our constructed \emph{DukeMTMC Group} and \emph{Road Group} datasets, respectively.}
	\vspace{-2mm}
\end{figure*}

\textbf{\bluet{Feature Notations.}}
The feature of an object $o \in \mathcal{O}_1$ in the fine granularity,
denoted by $\mathbf{f}_o = \mathbf{f}^l_o$,
is about the local appearance.
The feature of an object $o \in \mathcal{O}_2 \cup \mathcal{O}_3$ in the medium and coarse granularity,
denoted by $\mathbf{f}_o = [\mathbf{f}^l_o, \mathbf{f}^s_o]$, consists of two parts:
\blue{1) \emph{appearance}},
which is an aggregation \blue{of} features representing the local appearances of all people \red{within the subgroup},
and; \blue{2) \emph{spatial relation}}, \red{a single or aggregation of 
\blue{features}
$\mathbf{f}^s_o$ that describes the spatial layout of each edge linking two different individuals within this subgroup. \pink{Here, \emph{aggregation} indicates concatenating the feature of the same granularity and semantics, followed by performing t-SNE~\cite{maaten2008visualizing} for feature reduction.} \blue{The notation $[\cdot, \cdot]$ denotes a vector concatenation operation, which also applies to the rest of this paper.} Meanwhile, the global feature of object $o \in \mathcal{O}_g$, denoted by $\mbf{f}_o=\mbf{f}^g_o$, describes the appearance feature of the whole input \blue{group} image.}

\red{As shown in Fig.~\ref{fig:framework}, our framework is independent of the \blue{choice of} 
feature vectors used, which implies that we could exploit some conventional algorithms to extract hand-crafted features. On the other hand, we also \blue{intend} 
to exploit the strong ability of deep CNNs in extracting more representative and invariant \blue{features}. Therefore, we introduce two different pipelines to obtain these feature representations for usage in our group Re-ID framework. \blue{The first extracts} a combination of manually designed descriptors to \blue{encode both object appearances and layout relationship between objects} 
\blue{while the second is a new} integrated deep learning based method to extract features in a single forward pass.}

\subsection{Hand-crafted Feature Descriptors}\label{sec:hand}
\red{In this section, we briefly 
\blue{describe} 
the hand-crafted features utilized in this work.} Color and texture features~\cite{mirror} are used as
the \emph{appearance} part of the \blue{object.} 
\red{To be specific, for each single person input image (\blue{obtained} from bounding box $b_i$ and resized to unified resolution), we 
split it into $18$ equal-sized patches \blue{along the} vertical direction, and the RGB, HSV, YCbCr, LAB, YIQ color features and Gabor texture features are extracted \blue{from}
each patch. The output normalized histograms of these features are concatenated to form a final $8,024$-dimensional local \emph{appearance} descriptor $\mbf{f}^l_o$. Similarly, we also extract global features $\mbf{f}^g_o$ in the same manner but with the \blue{input image containing the entire group.}}

\red{As for the \emph{spatial relation} part,} we use the relative distance \& angle histograms among individuals in an object~\cite{harg} \red{to describe each edge between two people ($i$,$j$). For two bounding boxes $b_i$, $b_j$ within a subgroup, we first denote the relative position between their centers in polar coordinate $(\rho_{ij}, \theta_{ij})$, where $\rho_{ij}$ is the log-distance between the two centers and $\theta_{ij}$ is the corresponding orientation angle. \blue{The} 10-D log-distance histogram $L_{ij}$ and 9-D angle histogram $P_{ij}$ are constructed over uniform bins as follows:
}
\begin{align}\label{eq:distance-hist}
	L_{ij}(k) = \gassian(k-m;0,\sigma_L)
\end{align}
\begin{align}\label{eq:angle-hist}
P_{ij}(k) = \gassian(k-m_{ij};0, \sigma_P) +\gassian(k-m_{ij};\pm 9, \sigma_P)
\end{align}
\red{where $\gassian(x; \mu, \sigma)$ is a discrete Gaussian window parameterized by mean $\mu$ and variance $\sigma$, while $m_{ij}$ is the index of bin containing $\rho_{ij}$ or $\theta_{ij}$. \blue{Finally,} the two output histograms are combined
to form the descriptor that represents the $b_i$-$b_j$ edge:}
\begin{align}\label{eq:spatial-feature}
\mbf{f}^s_{(i,j)} = [L_{ij}, P_{ij}]
\end{align}
\subsection{Integrated CNN for Feature Extraction}\label{sec:cnn}

\red{Inspired by modern CNN based object detection frameworks ~\cite{Ren2015Faster}, which perform \blue{exceedingly} well in both recognition and localization tasks, we 
\blue{hypothesize}
that deep neural networks 
\blue{can be tailored}
 to handle both appearance and 
 \blue{structural layout of objects}
 in an integrated manner. As such,
 we propose a new multi-task network to \blue{jointly} extract both \emph{appearance} and \emph{spatial relation} features needed for our \blue{group Re-ID} framework.}

\red{The overview of the multiple-task CNN in our work is depicted in Fig. \ref{fig:cnn}. For each input image, we use the ResNet-34~\cite{He2015Deep} as the backbone structure for basic feature extraction. Post-processing includes two separate branches: The global branch, which is responsible for extracting features from the entire group image, and a local branch, which is utilized to handle \blue{individual objects and their relations to others}. \blue{With these branches processed in parallel,} 
we could obtain the multi-grained features simultanously.}

\subsubsection{Minibatch organization}
\red{We borrow the idea of using triplet loss for training~\cite{Hermans2017Trihard,Chen2017Beyond,Cheng2016Person} to learn representative mapping of images to an abstract feature space. Therefore, we organize the minibatch \blue{into triplets} 
where each training sample consists of three images: an anchor image $I_a$ from probe images, a positive image $I_p$ from the gallery which contains the same group as anchor, and a negative image $I_n$ which is randomly selected from training set \blue{that has} different group id from the anchor and positive images. }
	
\red{
\blue{Throughout this section,}
we shall adopt the same subscripts to denote anchor, positive and negative images for features or \blue{object}
sets. For example, $\mbf{f}^g_a$ denotes global feature of the anchor image and $\mathcal{O}_{1,a}$ denotes the set of person \blue{objects}
in the anchor image.}

\subsubsection{Global feature extraction}
The global branch receives feature maps from the final convolution layer of the 5-th stage of ResNet-34, which is denoted as S5 in Fig. \ref{fig:cnn}. \red{We then apply a simple global average pooling operation followed by a fully-connected layer as the global feature extraction module to obtain the corresponding global features $\mbf{f}_m^g$, $m \in \{a,p,n\}$.}

\subsubsection{Local feature extraction}
The local branch receives feature maps from the final convolution layer of the 4-th stage of ResNet-34 (denoted as S4 in Fig. \ref{fig:cnn}) and bounding box sets $\mathcal{B}_a, \mathcal{B}_p$ and $\mathcal{B}_n$ 
\blue{belonging to}
each image of the triplet. \red{The local feature extraction module applies ROI Pooling \cite{Ren2015Faster} on each feature map according to its bounding box and sends the output to a fully-connected layer. Hence, we obtain the intermediate local features $\hat{\mbf{f}}_{i,m}^l$ for each 
$i$-th individual person in the image of set} 
\blue{$m\in\{a, p, n\}$.}

Next, \blue{these features are further utilized in two sub-branches. First,} 
\red{a feature projection module takes the intermediate features as input and yields local appearance features \blue{by nonlinear projection that constrains outputs to (-1, 1)}:}
\begin{align}
	\mbf{{f}}_{i,m}^l = tanh(\mbf{W}_l\hat{\mbf{f}}_{i, m}^l)
\end{align}
\red{where $\mbf{W}_l$ is a learnable transformation matrix. 
\blue{Secondly, 
to model the spatial relation between two objects, i.e.} 
an edge linking two individuals ($i$,$j$), a feature fusion module is applied to fuse intermediate features with another learnable transformation matrix $\mbf{W}_s$ as follows}:
\begin{align}
\mbf{f}_{(i,j),m}^s = tanh(\mbf{W}_s[\hat{\mbf{f}}_{i,m}^l, \hat{\mbf{f}}_{j,m}^l])
\end{align}

\subsubsection{Loss function}
\red{To regularize the global and local output features from different branches, we design three different types of losses to train our multi-task CNN. For the global \blue{appearance features,}
\blue{we intend to learn an embedding space where} 
the anchor sample will be closer to the positive sample than the negative sample. \blue{For this}, we obtain a group-wise training loss by utilizing the triplet loss:}
\begin{align}\label{eq:global-loss}
L_g =[d_2(\mbf{f}^g_a, \mbf{f}^g_p) - d_2( \mbf{f}^g_a,\mbf{f}^g_n) + \lambda_g]_+
\end{align}
\red{where $[\cdot]_+$ is the ReLU operator $\max(0,\cdot)$, $d_2(\cdot,\cdot)$ denotes the L2-norm distance between the \blue{two feature} vectors, and $\lambda_g$ is a hyper-parameter to control the margin size.}

\red{For the supervision of individual appearance features, we could simply apply a triplet loss over the matched and unmatched pairs, similar to that in Eq. \ref{eq:global-loss}. However, since the number of matched individual pairs between anchor and positive images is much lesser than that of unmatched ones, the loss might well be \blue{dominated} by unmatched pairs. Inspired by the Trihard loss \cite{Hermans2017Trihard}, we impose the hard negative mining strategy to the standard triplet loss,
\blue{much akin to}
a k-nearest negative neighbors manner:}
\begin{align}\label{eq:fine-grained-loss}
	L_a = \frac{1}{\vert \mathcal{O}_{1,a} \vert} \sum_{i \in \mathcal{O}_{1,a}}\left[d_2(\mbf{f}^l_{i,a},\mbf{f}^l_{i',p})-d_{neg}(i)+\lambda_l\right]_+
\end{align} 
\red{where $i' \in \mathcal{O}_{1,p}$ is the individual person in the positive image who matches exactly to the $i$-th person in the anchor image. Note that if person $i$ in anchor does not match any individual in positive image, we set the $d_2(\cdot,\cdot)$ term in Eq. \ref{eq:fine-grained-loss} to be zero. The hyper-parameter $\lambda_l$ controls the margin size of local appearance features. The term $d_{neg}(i)$ is the average 
distance between person $i$ in anchor image and persons in set $\mathcal{K}= \mathcal{K}_p \cup \mathcal{K}_n$, which \blue{carries the intuition} 
of $k$-nearest unmatched individuals from other images in the feature space:}
\begin{align}
	d_{neg}(i)=\frac{1}{|\mathcal{K}|}\left(\sum_{j\in \mathcal{K}_p}d_2(\mbf{f}^l_{i,a},\mbf{f}^l_{j,p})+\sum_{j \in \mathcal{K}_n}d_2(\mbf{f}^l_{i,a},\mbf{f}^l_{j,p})\right)
\end{align}
\red{where $\mathcal{K}_p$ and $\mathcal{K}_n$ denotes the collection of $k$ nearest unmatched individuals from positive and negative images respectively.}

\red{In the work of \cite{Ren2015Faster}, deep neural networks can precisely predict the relative \blue{offset}
between two bounding boxes. Based on this observation, we design a regression loss to supervise the learning of spatial relation features. Given the bounding boxes, $b_{i,m}, b_{j,m} \in \mathcal{B}_m$ (with $m \in \{a,p,n\}$ as mentioned before) of two individuals ($i$,$j$) in an image and the feature $\mbf{f}_{(i,j),m}^s$ representing the edge between them, we apply linear regression with learnable parameter $W_r$ to predict the normalized spatial transition:}
\begin{align}
	[\hat{\delta}^x_{(i,j),m}, \hat{\delta}^y_{(i,j),m}]^T = W_r \mathbf{f}^s_{(i,j),m}
\end{align}
\red{Suppose the bounding box is in the form of $b_{m, i}=(x_{i,m}, y_{i,m}, w_{i,m}, h_{i,m})$ as defined in \cite{Ren2015Faster}, we denote the ground truth transition as:
}
\begin{align}
	& \delta^x_{(i,j),m} = \frac{x_{j,m}-x_{i,m}}{w_{i,m}} \\
	& \delta^y_{(i,j),m} = \frac{y_{j,m}-y_{i,m}}{h_{i,m}}
\end{align}
\red{Hence, the localization loss over spatial relation features is:}
\begin{align}\label{eq:spatial loss}
	L_s = \frac{1}{P} \sum_{m\in\{a, p, n\}}\sum_{(i,j) \in \mathcal{O}_{2,m}} \sum_{t\in \{x,y\}} S\left(\hat{\delta}^t_{(i,j),m}, \delta^t_{(i,j),m}\right)
\end{align}
\red{where $P=\sum_{m \in \{a,p,n\}}|\mathcal{O}_{2,m}|$ is the sum of total number of bounding box     pairs within each image of the triplet. $S(\cdot,\cdot)$ denotes the smooth L1 loss used also in \cite{Ren2015Faster}. Finally we combined the losses in Eq. \ref{eq:global-loss}, Eq. \ref{eq:fine-grained-loss} and Eq. \ref{eq:spatial loss} to obtain the final objective function for our multi-task CNN framework:}
\begin{align}
	L = L_g + \lambda_1 L_a +\lambda_2 L_s
\end{align}
where $\lambda_1$, $\lambda_2$ are balancing factors. \red{After the training phase, this multi-task network is used to extract deep multi-grained features in a single forward pass. \blue{Aggregation of these local features is performed} to attain features of coarse and medium granularities for the \blue{subsequent} matching step.} \redt{In our experiments, we set $\lambda_g$ and $\lambda_l$ as $2.0$, and the balancing factors $\lambda_1$ and $\lambda_2$ as $1.0$ during training.} 

\section{Importance Weighting \label{subsection:feature extract2}}
We introduce an importance weight $\alpha_o$ for each object $o$
(except the global-grained object) to
indicate the object's discriminativity and reliability inside the group image
for group person matching.
The importance weighting scheme is partially inspired by
but different from
the saliency-learning methods~\cite{zhaorui2,icip16}
for differentiating~\emph{patch} reliabilities in person re-identification:
(i) Our scheme aims to weight each granularity object
rather than patches;
(ii) Our scheme dynamically adjusts the importance weights
in an iterative manner, by using the intermediate matching results at each iteration.
(cf. Fig.~\ref{fig:framework}).

\subsection{Fine-grained Object}
The importance weight ($\alpha_i$)
for each individual person $i$
in the probe image $I$
consists of
two components:
static weight,
which is only dependent on the group image,
and dynamic weight,
which is dynamically updated
according to the intermediate matching results
with the gallery group images,
from another camera in our approach.
The formulation of this weight is given as follows,
\begin{align}
\alpha_i=
t_1(i,\mathcal{G}_{\backslash i})
+
s(i,\mathcal{M}_i)+
p(\mathcal{M}_i,\mathcal{M}_{\mathcal{G}_{\backslash i}}),
\label{eq:importance}
\end{align}
where the first term is the static weight,
and the second and third terms form the dynamic weight.

\vspace{.1cm}
\noindent\textbf{Static weight.}
The static weight $t_1(i,\mathcal{G}_{\backslash i})$,
where $\mathcal{G}_{\backslash i} = \mathcal{G} - \{i\}$
denotes the set of other individual people in $\mathcal{G}$,
is used to describe the stability.
It is computed as follows,
\begin{align}
t_1(i,\mathcal{G}_{\backslash i})
= \sum_{i' \in \mathcal{G}_{\backslash i}} \frac{\rho_i}{\rho_{i'}},
\label{eq:stability}
\end{align}
where $\rho_i$ is the local density around person $i$ in group $\mathcal{G}$. {It reflects the density of people in a neighborhood around $i$,} which is computed by following~\cite{lof}.

\begin{figure}[t]
	\centering
	\vspace{-1mm}
	\subfloat[]{\includegraphics[width=0.35\textwidth]{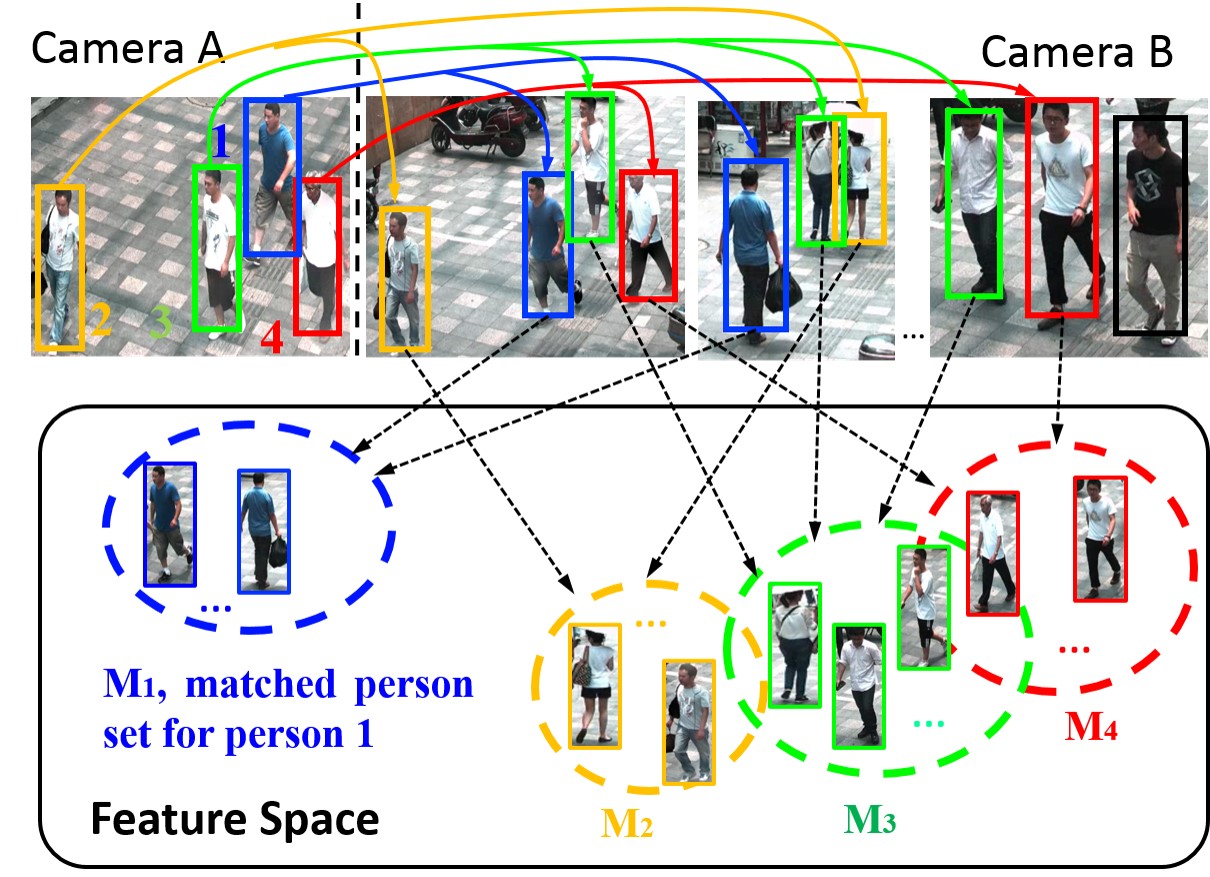}    \label{fig:sal_a}}
	\\
	\subfloat[]{\includegraphics[width=0.4\textwidth]{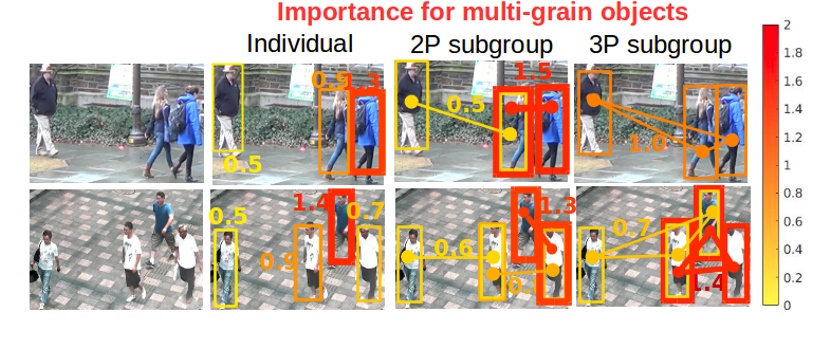}\label{fig:sal_b}}
	\caption{(a) Illustration of matched-people sets and their distributions in the feature space (The color solid arrows indicate the one-to-one mapping results between individuals. People circled by the same-color rectangles in camera $B$ are matched to the same person in $A$, and belong to the same matched-people set). (b) The derived importance weights for multi-grained objects (individuals, $2$-people subgroups, $3$-people subgroups) in two example group images. Note: the importance weights for some $2$-people/$3$-people subgroups are not displayed in order for a cleaner illustration. (Best viewed in color)}
\end{figure}

{By Eq.~\ref{eq:stability}, the static weight $t_1$ is mainly obtained by evaluating the relative local density ratios between person $i$ and his/her peer group members ${i{'}}$ in $\mathcal{G}$. If the local density around $i$ is larger than the density around his/her peer group members ${i{'}}$, the stability of $i$ is increased, indicating that $i$ is located in the \emph{center} region of group $\mathcal{G}$ and should be a more reliable member in group Re-ID (cf. person $1$ in Fig.~\ref{fig:sal_a}). On the contrary, when $i$'s local density is smaller than his/her peer group members, a small stability value will be assigned, indicating that $i$ is located in the outlier region of the group and is less reliable (cf. person $2$ in Fig.~\ref{fig:sal_a}).
}

\vspace{.1cm}
\noindent\textbf{Dynamic weight.}
The dynamic weight
$s(i,\mathcal{M}_i)+
p(\mathcal{M}_i,\mathcal{M}_{\mathcal{G}_{\backslash i}})$
consists of two parts:
the saliency term $s(i,\mathcal{M}_i)$
and the purity term $p(\mathcal{M}_i,\mathcal{M}_{\mathcal{G}_{\backslash i}})$,
where $\mathcal{M}_i$ is the set of matches from the gallery group images,
and $\mathcal{M}_{\mathcal{G}_{\backslash i}}$ is the set of the matches for all people except $i$
in the probe image,
$\mathcal{M}_{\mathcal{G}_{\backslash i}} = \{ \mathcal{M}_{i'} | i' \notin \mathcal{G} \}$.
The sets of matches are illustrated in Fig.~\ref{fig:sal_a}.

The saliency term is computed as
\begin{align}
s(i, \mathcal{M}_i) = \lambda_s \frac{d_f(\mathbf{f}_i, \mathbf{f}_{\mathcal{M}_i})}{|\mathcal{M}_i|}.
\label{eq:saliency}
\end{align}
Here $d_f(\cdot)$ is the Euclidean distance between features.
$|\mathcal{M}_i|$ is the cardinality of $\mathcal{M}_i$.
$\mathbf{f}_{\mathcal{M}_i}$ is the feature describing the set of matches $\mathcal{M}_i$,
and we use the feature of the $\frac{1}{2}|\mathcal{M}_i|$th nearest neighbor of $i$ in $\mathcal{M}_i$
as done in \cite{zhaorui1,zhaorui2}.
$\lambda_{s}$ is an \redt{ adaptive normalization factor} to normalize the range of ${s}$ to be within $0$ to $1$, \redt{ which is computed \bluet{as follows:} 
}
\begin{equation}
    \lambda_s = \frac{1}{\sum_i{d_f(\mathbf{f}_i, \mathbf{f}_{\mathcal{M}_i})/|\mathcal{M}_i|}}\label{eq:ls}
\end{equation}
For simplicity, the other normalization factors in the rest of this paper are calculated in the similar way as $\lambda_s$.

{
	According to Eq.~\ref{eq:saliency}, if the appearance of an individual person $i$ is discriminative, a large portion of individuals in $i$'s matched set $\mathcal{M}_i$ are visually dissimilar to $i$. This leads to a large $d_f(\mathbf{f}_i, \mathbf{f}_{\mathcal{M}_i})$ and a large saliency value~\cite{zhaorui1,zhaorui2} (cf. person $1$ in Fig.~\ref{fig:sal_a}). }{ Moreover, due to the variation of group members in group Re-ID, each individual person may have different number of matched people in his/her $\mathcal{M}_i$. Therefore, we further introduce $|\mathcal{M}_i|$ in Eq.~\ref{eq:saliency}, such that person with fewer matched people can indicate more discriminative appearance.}

The purity term is computed as
\begin{align}
p(\mathcal{M}_i,\mathcal{M}_{\mathcal{G}_{\backslash i}})
= \sum_{i' \in \mathcal{G}_{\backslash i}}
\lambda_{p} d_m (\mathcal{M}_i, \mathcal{M}_{i'}),
\label{eq:purity}
\end{align}
where $d_m(\cdot)$ is the Wasserstein-$1$ distance \cite{rubner2000earth},
a measure to evaluate the dissimilarity between two feature sets.
{$\lambda_{p}$ is calculated in the same way as $\lambda_{s}$ in Eq.~\ref{eq:saliency}.}

{According to Eq.~\ref{eq:purity} and Fig.~\ref{fig:sal_a}, the purity measurement reflects the relative appearance uniqueness of person $i$ inside group $\mathcal{G}$. If $i$ has similar appearance features as other group members in $\mathcal{G}$, their matched people in camera $B$ should also be visually similar and located close to each other in the feature space (cf. $\mathcal{M}_3$ and $\mathcal{M}_4$ in Fig.~\ref{fig:sal_a}), resulting in a small purity value. On the other hand, if a person includes \emph{unique} appearance features in $\mathcal{G}$, his/her matched people in camera $B$ should have larger feature distances to those of the other members in $\mathcal{G}$, and lead to a large purity value (cf. $\mathcal{M}_1$ in Fig.~\ref{fig:sal_a}).
}

\subsection{Medium and Coarse grained Objects}
The importance weight $\alpha_{i_1i_2}$  of a medium-grained object $(i_1, i_2)$ is computed as:
\begin{align}
\label{equation:importance2}
\alpha_{i_1i_2}=\alpha_{i_1}+\alpha_{i_2}+{t}_2(i_1,i_2).
\end{align}
Here, ${t}_2(i_1,i_2)$ is the stability measure of the sub-group $(i_1,i_2)$.
A two-people sub-group is thought
more stable if its members are spatially closer to each other.
Thus, we simply compute ${t}_2$ by the inverse of spatial distance between $i_1$ and $i_2$.

The importance weight $\alpha_{i_1i_2i_3}$  of a coarse-grained object $(i_1, i_2, i_3)$
is computed as:
\begin{align}
\label{equation:importance3}
\alpha_{i_1i_2i_3}=\alpha_{i_1i_2}+\alpha_{i_2i_3}
+\alpha_{i_1i_3}+{t}_3(i_1,i_2,i_3).
\end{align}
Here, $\alpha_{i_1i_2}$ is the importance of a two-people pair in  $(i_1,i_2,i_3)$ (cf. Eq.~\ref{equation:importance2}).
${t}_3$ is the stability of a three-people subgroup.
We assume equilateral triangle as the most stable structure for three-people subgroups
and model ${t}_3$ by evaluating its similarity \red{to a equilateral triangle according to Eq. \ref{eq:triangle sim}, where $\theta_k, k\in \{1,2,3\}$ denote the angles of the triangle constructed by coarse sub-group $(i_1, i_2, i_3)$.}

\begin{align}\label{eq:triangle sim}
    t_3(i_1, i_2, i_3) = \exp{\left(-2*\sum_{k=1}^{3}{\vert\sin{\theta_k}-\sin{\frac{\pi}{3}}}\vert\right)}
\end{align}

Fig.~\ref{fig:sal_b} shows the importance weights of some groups. From Fig.~\ref{fig:sal_b}, we can see that our process can effectively set larger weights on objects \red{with stronger characterization ablility to represent the entire group}.

\subsection{Iterative Update}\label{iterativeprocess}
We utilize an iterative process which updates the importance weights and group-wise matching results iteratively.
We initialize the dynamic weights for all objects by $1$
and compute the optimal matching through multi-order matching (cf. Sec.~\ref{section:matching})
to obtain an initial matching result:
$\mathcal{M}_1, \mathcal{M}_2, \dots, \mathcal{M}_N$.
This matching result
is used to update the dynamic importance weights.
This procedure is repeated until the importance weights become converged or the maximum iteration is reached.
Although the exact convergence of our iterative process is difficult to analyze
due to the inclusion of multi-order matching,
in our experiments we confirm that most importance weights become stable within $5$ iterations, which implies the reliability of our approach.

\section{Multi-order Matching}\label{section:matching}
Given a probe image $I_p$ and a gallery image $I_g$,
our goal is to compute the matching score
between the two groups of people.
Suppose that there are $N_p$ people in the probe image $I_p$
and $N_g$ people in the gallery image $I_g$.
The goal of the multi-order matching process
aims to find: (1) an
optimal one-to-one mapping,
$\mathcal{C} = \{(i, j)|~\forall (i, j), (i{'}, j{'}), i \neq i{'}, j \neq j{'}\}$,
where $(i, j)$ ($=c_{ij}$)
denotes a match between the $i$th person from the probe image
and the $j$th person from the gallery image,
and (2) the maximum matching score.

\redt{Since a group is characterized by multiple granularities, it is natural to measure the similarity \bluet{across} 
different granularities to find the optimal match.} With this consideration, The objective function of our matching process is formulated
with multi-order potentials:
\begin{align}
\mathcal{Q}(\mathcal{C}) =
&\mathcal{P}_1(\mathcal{C}) + \mathcal{P}_2(\mathcal{C}) + \mathcal{P}_3(\mathcal{C}) + \mathcal{P}_g(\mathcal{C}) \notag \\
&+ \sum_{r \neq l, r,l \in \{ 1, 2, 3, g\}}
\mathcal{P}_{rl}(\mathcal{C}), \label{eq:objective}
\end{align}
where $\mathcal{P}_1(\mathcal{C})$,
$\mathcal{P}_2(\mathcal{C})$,
$\mathcal{P}_3(\mathcal{C})$,
and $\mathcal{P}_g(\mathcal{C})$
are the first-order, second-order, third-order, and global potentials,
evaluating the matching quality over
each subgroup of people,
and $\mathcal{P}_{rl}(\mathcal{C})$ is the inter-order potential.

\subsection{Multi-order Potentials}
\noindent\textbf{First order potential.}
$\mathcal{P}_1(\mathcal{C})$ is used to model the matching scores over individual people. It is calculated by the sum
of the matching scores
of all the individual matches in $\mathcal{C}$:
\begin{align}
\mathcal{P}_1(\mathcal{C})
=~ \sum_{c_{ij} \in \mathcal{C}} m_1(c_{ij})=~ \sum_{c_{ij} \in \mathcal{C}} w_1(\mathbf{f}_i, \alpha_i, \mathbf{f}_j, \alpha_j)
\label{eq:weight11}
\end{align}
where {$\mathbf{f}_i$, $\alpha_i$} and {$\mathbf{f}_j$, $\alpha_j$} are the feature vector and importance weight for probe-image person $i$ and gallery-image person $j$, respectively (cf. Eq.~\ref{eq:importance}). $m_1(c_{ij})=w_1(\mathbf{f}_i, \alpha_i, \mathbf{f}_j, \alpha_j)$ is the matching score for match $c_{ij}=(i,j)$, calculated by:
\begin{align}
m_1(c_{ij})=w_1(\mathbf{f}_i, \alpha_i, \mathbf{f}_j, \alpha_j) =\lambda_{w_1}\frac{\psi(\alpha_i,\alpha_j)}{d_\mathbf{f}(\mathbf{f}_i,\mathbf{f}_j)}
\label{eq:weight2}\end{align}
where $\psi(\alpha_i,\alpha_j)=\frac{\alpha_i+\alpha_j}{1+|\alpha_i-\alpha_j|}$ is the fused importance weight, which will have large value if the importance weights of $\alpha_i$ and $\alpha_j$ are both large and close to each other. $d_\mathbf{f}(\cdot)$ is the Euclidean distance and $\lambda_{w_1}$ is the normalization constant for the first-order potential.

By Eq.~\ref{eq:weight2}, the matching score $m_1(c_{ij})$ is computed by the importance-weighted feature similarity $w_1(\mathbf{f}_i, \alpha_i, \mathbf{f}_j, \alpha_j)$ between the matched individuals $i$ and $j$.

\vspace{.1cm}
\noindent\textbf{Second order potential.}
$\mathcal{P}_2(\mathcal{C})$ is used
to model the matching scores over two-people subgroups:
\begin{align}
\mathcal{P}_2(\mathcal{C})
=~& \sum_{c_{i_1j_1}, c_{i_2j_2} \in \mathcal{C}} m_2(c_{i_1j_1}, c_{i_2j_2}) \notag\\
=~& \sum_{c_{i_1j_1}, c_{i_2j_2} \in \mathcal{C}} w_2(\mathbf{f}_{i_1i_2}, \alpha_{i_1i_2},
\mathbf{f}_{j_1j_2}, \alpha_{j_1j_2}).
\end{align}
where {$\mathbf{f}_{i_1i_2}$, $\alpha_{i_1i_2}$} and {$\mathbf{f}_{j_1j_2}$, $\alpha_{j_1j_2}$} are the feature vector and importance weight for probe-image subgroup $(i_1,i_2)$ and gallery-image subgroup $(j_1,j_2)$, respectively (cf. Eq.~\ref{equation:importance2}). $m_2(c_{i_1j_1}, c_{i_2j_2})=w_2(\mathbf{f}_{i_1i_2}, \alpha_{i_1i_2},
\mathbf{f}_{j_1j_2}, \alpha_{j_1j_2})$ is the second order match score between two-people subgroups $(i_1,i_2)$ and $(j_1,j_2)$, which is calculated in a similar way as Eq.~\ref{eq:weight2}:
\begin{align}
w_2(\mathbf{f}_{i_1i_2}, \alpha_{i_1i_2},
\mathbf{f}_{j_1j_2}, \alpha_{j_1j_2}) =\lambda_{w_2}\frac{\psi(\alpha_{i_1i_2},\alpha_{j_1j_2})}{d_\mathbf{f}(\mathbf{f}_{i_1i_2},\mathbf{f}_{j_1j_2})}
\label{eq:weight3}\end{align}

\vspace{.1cm}
\noindent\textbf{Third order potential.}
$\mathcal{P}_3(\mathcal{C})$ is used
to model the matching scores over three-people subgroups:
\begin{align}
\mathcal{P}_3(\mathcal{C})
=~& \quad \sum_{\mathclap{c_{i_1j_1}, c_{i_2j_2}, c_{i_3j_3} \in \mathcal{C}}} \quad m_3(c_{i_1j_1}, c_{i_2j_2}, c_{i_3j_3}) \notag\\
=~& \quad \sum_{\mathclap{c_{i_1j_1}, c_{i_2j_2}, c_{i_3j_3} \in \mathcal{C}}} \quad w_3(\mathbf{f}_{i_1i_2i_3}, \alpha_{i_1i_2i_3},
\mathbf{f}_{j_1j_2j_3}, \alpha_{j_1j_2j_3}).
\end{align}
where the term $w_3(\mathbf{f}_{i_1i_2i_3}, \alpha_{i_1i_2i_3},
\mathbf{f}_{j_1j_2j_3}, \alpha_{j_1j_2j_3})=m_3(c_{i_1j_1}, c_{i_2j_2}, c_{i_3j_3})$ is the third order match score between three-people subgroups $(i_1,i_2,i_3)$ and $(j_1,j_2,j_3)$. It is calculated in the same way as Eq.~\ref{eq:weight3}.

\vspace{.1cm}
\noindent\textbf{Global potential.}
The global potential is calculated by the global matching score between probe and gallery images $I_p$ and $I_g$:
\begin{align}
\mathcal{P}_g(\mathcal{C})
=~& \sum_{\mathcal{C}} m_g(c_{i_1j_1}, c_{i_2j_2}, \dots, c_{i_{N_p}j_{N_q}}) \notag\\
=~& w_g(\mathbf{f}_{p}, \alpha_{p}, \mathbf{f}_{g}, \alpha_{g})
\label{eq:weight4}
\end{align}
where $\mathbf{f}_{p}$ and $\mathbf{f}_{g}$ are the global feature vectors for the entire group images $I_p$ and $I_g$. $\alpha_{p}=\alpha_{g}=1$ are the importance weights for global objects. In this paper, we simply use the global feature similarity as the global matching score, as: $w_g(\mathbf{f}_{p}, \alpha_{p}, \mathbf{f}_{g}, \alpha_{g})=\frac{1}{d_\mathbf{f}(\mathbf{f}_{p},\mathbf{f}_{g})}$.

\vspace{.1cm}
\noindent\textbf{Inter-order potential.} Since each match $c_{ij}$ is described by potentials in multiple orders (cf. Eqs.~\ref{eq:weight11}-\ref{eq:weight4}), { we also introduce inter-order potentials to properly combine these multi-order potential information.} Specifically, the inter-order potential between orders $r, l \in \{1, 2, 3, g\}$ is calculated by:

\begin{align}
\mathcal{P}_{rl}(\mathcal{C})
=~& \sum_{c_{ij} \in \mathcal{C}} m_{rl} (c_{ij}, r, l)
\end{align}
where $m_{rl} (c_{ij}, r, l)$ is the {inter-order correlation} for match $c_{ij}$. It is calculated by:

\begin{align}
m_{rl} (c_{ij}, r, l&)=\frac{{\overline{m}}_{r}(c_{ij}, r)
	+\overline{m}_{l}(c_{ij}, l)}{1+|\overline{m}_{r}(c_{ij}, r)-\overline{m}_{l}(c_{ij}, l)|} \notag \\
\text{for }~ {\overline{m}}_{k}(c_{ij}, k)=&\lambda_{{k}} \sum_{~c_{i^{'}_1j^{'}_1}=c_{ij}} m_{k}(c_{i^{'}_1j^{'}_1}...c_{i^{'}_{k}j^{'}_{k}})\label{eq:weight5}
\end{align}
where $\lambda_{k}$ is the normalization constant for order $k$, $m_{k}(c_{i^{'}_1j^{'}_1}...c_{i^{'}_{k}j^{'}_{k}})$ is the intra-order match score in order $k$ (as in Eqs.~\ref{eq:weight2} and \ref{eq:weight3}).
From Eq.~\ref{eq:weight5}, if a match $c_{ij}$ creates large and similar intra-order match scores in both the $r$-th and $l$-th order, it will be considered as more valuable and reliable, and thus will have larger inter-level potentials.

\subsection{Optimization}

The objective function in Eq.~\ref{eq:objective} properly integrates the information of multi-grained objects. Thus, by maximizing Eq.~\ref{eq:objective}, we are able to obtain the optimal one-to-one mapping result among individuals in probe and gallery groups.

\begin{figure}[tb]
	\centering
	\vspace{0mm}
	\includegraphics[width=0.46\textwidth,height=0.21\textwidth]{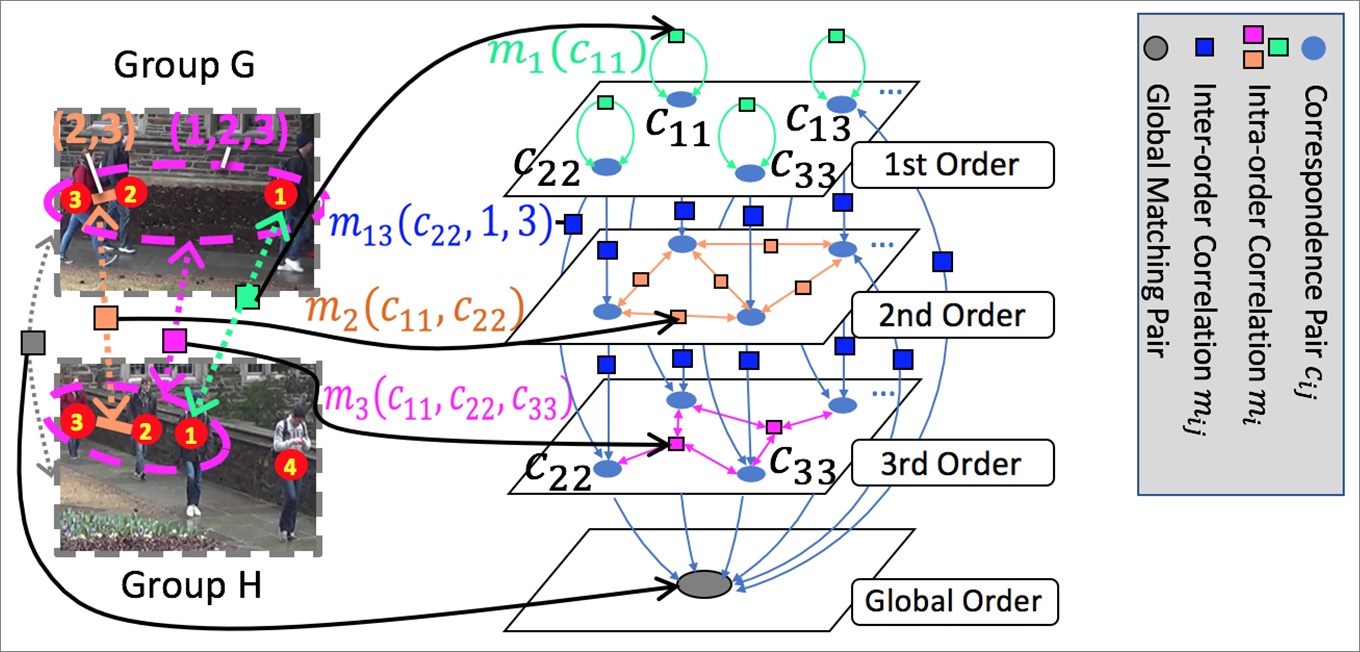}
	\caption{Illustration of multi-order association graph. Left: A cross-view group pair being matched; Right: The multi-order association graph constructed for the group pair. (Best viewed in color)}
	\label{fig:matching}
\end{figure}

To solve the multi-order matching problem in Eq.~\ref{eq:objective}, we construct a multi-order association graph to incorporate all candidate matches \& multi-order potentials in the objective function, as in Fig.~\ref{fig:matching}. In Fig.~\ref{fig:matching}, each layer includes all candidate matches $c_{ij}$ (the circular nodes) and their corresponding intra-order matching scores $m_k$ (the rectangular nodes in green, orange, or pink), which models the intra-order potentials in a specific order. Besides, the blue rectangular nodes linking circular nodes in different layers represent the inter-order correlations $m_{rl} (c_{ij}, r, l)$. They model the inter-order potentials between different orders.

With this association graph, we are able to solve Eq.~\ref{eq:objective} by adopting general-purpose hyper-graph matching solvers \cite{yan2017adaptive,RRWHM}. Specifically, we first initialize a mapping probability for each candidate match in the associate graph, and then apply reweighted random walk~\cite{RRWHM} to update these mapping probabilities via the inter/intra-order links and potential weights in the association graph. Finally, the mapping probabilities in all layers in the association graph are combined to obtain the optimal one-to-one mapping result from the candidate matches~\cite{yan2017adaptive}.

\subsection{Fused Matching Score}

\begin{figure}
	\centering
	\centering
	\vspace{0mm}
	\subfloat[]{\includegraphics[width=0.21\textwidth,height=0.11\textwidth]{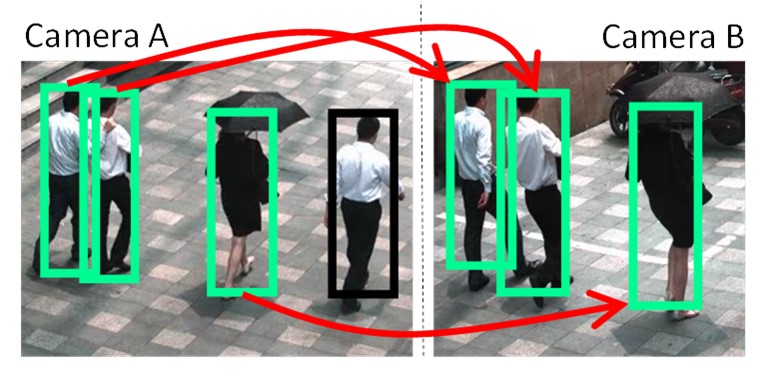}\label{fig:different matching score a}}
	\vspace{1mm}	
	{\vrule width0.6pt}
	\subfloat[]{\includegraphics[width=0.25\textwidth,height=0.11\textwidth]{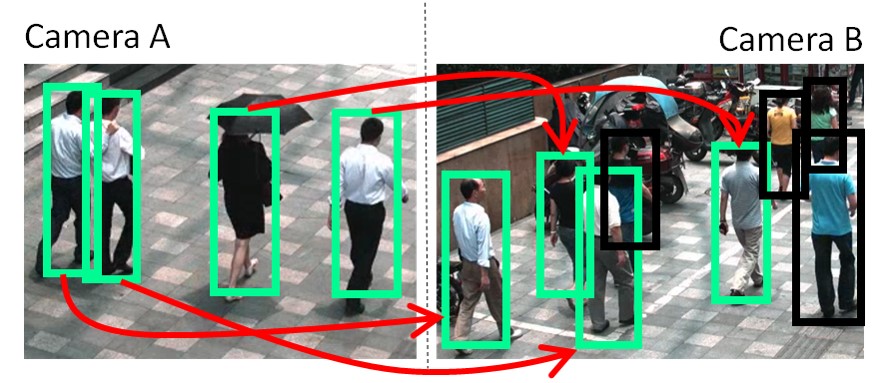}\label{fig:different matching score b}}
	\caption{Illustration of the unmatched term in Eq.~\ref{equation:score}. (a) is a true match pair and (b) is a false match pair. Green and black rectangles show matched and unmatched individuals, respectively. Since the right group in (b) includes more individuals, we can find more matched pairs. This may misleadingly result in a high similarity score. However, if considering the large number of unmatched people in (b), the matching score of (b) ought to be properly reduced.}
	\label{fig:unmatched}
\end{figure}

After obtaining one-to-one mapping between individual people in two groups, we are able to calculate matching scores accordingly. In order to obtain a more reliable matching score, we introduce a fused scheme by integrating the information of both matched and unmatched objects:
\vspace{-0.5mm}
\begin{xiaowuhao}\begin{align}
	\label{equation:score}
	&S(I_p,I_g)=\sum_{k} \quad \sum_{\mathclap{~~(i_1..i_k)\in \mathcal{\mathcal{R}}_p}} \frac{w_k(\mathbf{f}_{i_1..i_k}, \alpha_{i_1..i_k},
		\mathbf{f}_{M({i_1..i_k})}, \alpha_{M({i_1..i_k})})}{{|\mathcal{\mathcal{R}}_p|}}\notag  \\
	&-\lambda_{r} \cdot \sum_{k} \left(\quad \ \ \ \sum_{\mathclap{(i_1..i_k)\in\overline{\mathcal{\mathcal{R}}}_p}} \qquad \ \frac{a_{i_1..i_k}}{|\overline{\mathcal{\mathcal{R}}}_p|} + \sum_{{(j_1..j_k)\in\overline{\mathcal{\mathcal{R}}}_g}} \ \frac{a_{j_1..j_k}}{|\overline{\mathcal{\mathcal{R}}}_g|} \right)
	\end{align}\end{xiaowuhao}
\noindent where $({i_1,..,i_k})$ is a person/subgroup in probe group image $I_p$, $M({i_1..i_k})$ is its one-to-one matched person/subgroup in gallery image $I_g$. $w_k(\cdot)$ is the similarity matching score between $({i_1,..,i_k})$ and $M({i_1..i_k})$, as in Eqs.~\ref{eq:weight2} and \ref{eq:weight3}. $\alpha$ is the importance weight. $\lambda_{r}$=$0.5$ is a balancing factor. $\mathcal{\mathcal{R}}_p$ and $\mathcal{\mathcal{R}}_g$ are the sets of reliably matched objects in groups $I_p$ and $I_g$, and $\overline{\mathcal{\mathcal{R}}}_p$ and $\overline{\mathcal{\mathcal{R}}}_g$ are the unmatched object sets. The matched object pairs that maximizing objective function~\ref{eq:objective} are taken as the reliably matched objects, and put into $\mathcal{\mathcal{R}}_p$ and $\mathcal{\mathcal{R}}_g$. The remaining unmatched or less similar objects are put into $\overline{\mathcal{\mathcal{R}}}_p$ and $\overline{\mathcal{\mathcal{R}}}_g$.

From Eq.~\ref{equation:score}, our fused scheme integrates four granularities (i.e., $k=1,2,3,g$) to compute the group-wise matching score. Inside each granularity, \red{the matched pairs that maximize Eq. ~\ref{eq:objective} are used} to compute the similarity (the first term in Eq.~\ref{equation:score}), so as to reduce the interference of confusing or mismatched people/people subgroups. Meanwhile, we introduce an unmatched term evaluating the importance of unmatched objects (the second term in Eq.~\ref{equation:score}). As such, we can properly avoid misleadingly high matching scores in false group pairs (as in Fig.~\ref{fig:unmatched}) and obtain a more reliable result.


\begin{table}[t]
    \centering
    \caption{\bluet{Statistical} Analysis of Datasets}
    \begin{tabular}{lccc}\hline
        \multicolumn{1}{r}{Datasets} & \emph{i-LID MCTS} & \emph{Road Group} & \emph{DukeMTMC Group} \\ \hline
        Average Person & 2.313 & 3.812 & 3.392 \\
        Member Change & 0.187 & 0.451 & 0.832 \\
        Dispersity & 0.317&0.376&0.407 \\
        Occlusion Pairs & 1.021 & 1.775 & 2.001 \\ 
        \hline
    \end{tabular}
    \label{tab:dataset}
\end{table}

\section{Experimental Results\label{section:experimental evaluation}}

We perform experiments on three datasets: (1) the \blue{publicly available} \emph{i-LID MCTS} dataset \cite{zheng} which contains $274$ group images for $64$ groups; (2) our \blue{newly} constructed \emph{DukeMTMC Group} dataset which includes $177$ group image pairs extracted from the $8$-camera-view DukeMTMC dataset \cite{2016MTMC}; (3) our \blue{newly collected} 
\emph{Road Group} dataset which consists of $162$ group pairs taken from a $2$-camera \blue{crowded}
road scene\footnote{Dataset and source code will be available at \url{http://min.sjtu.edu.cn/lwydemo/GroupReID.html}.}.

To construct the \emph{Road Group} dataset, we use~\cite{groupdetection} to automatically \pink{identify groups from key frames that were extracted at equi-intervals of 50 frames. Then,}
\pink{the group image pairs are randomly selected from sets according to different group sizes and occlusion variations.} 
We define two cross-view images \blue{from different cameras} as belonging to the same group when they have more than $60\%$ members in common.
\begin{figure}[t]
	\centering
	
\end{figure}

Some example of groups from the three datasets are shown in Fig.~\ref{fig:datasets}, \blue{showing diverse challenging conditions across cameras.} \redt{We also 
\bluet{provide a statistical} analysis of the three datasets 
in Table~\ref{tab:dataset}.} `Average Person' denotes \bluet{the average number of} individuals per group image, while `Member Change' denotes the average difference in group size 
for each pair of group images. `Dispersity' is \bluet{measured} by averaging the normalized distance between each \bluet{individual member} to its group centroid. \bluet{This measure computes the sparsity (or compactness) of the group, which indirectly indicates the proneness} to layout change. 
`Occlusion Pairs' denotes the \bluet{average} number of individual pairs that occlude each other within an image. Note that \bluet{despite} the \emph{i-LID MCTS} dataset \bluet{having smaller and more compact groups, it} suffers from low image quality and large illumination \bluet{changes}. \bluet{Meanwhile,} the new datasets, \emph{DukeMTMC Group} and \emph{Road Group} are both \blue{plagued with} severe object occlusions, and large layout and group member variations \redt{due to larger \bluet{group sizes}}.

\subsection{Experiment Setup}

To \blue{provide} a fair comparison with other methods, we follow the evaluation protocol in~\cite{icip16,Gray2008Viewpoint} \red{by partitioning the datasets by half into training set and 
validation set. The final results are obtained by averaging the performances on the validation set over $5$ random splits. We report our results \blue{using}
} Cumulated Matching Characteristic (CMC)~\cite{shenyang}, \blue{which is able to measure rank-$k$ correct match rates.}

We use a ResNet-34 model as our model feature extractor. We initialize the learning rate at $10^{-4}$ and divide it by a factor of $10$ every $15$ epochs, \blue{stopping at the maximum epoch of $40$. 
Our network is trained with a SGD solver (weight decay of $10^{-4}$) on an NVIDIA GeForce GTX 1080 GPU.
\blue{Contrary to the work of}
\cite{Ren2015Faster}, we employ a 
\blue{non-conventional}
bin distribution of size $6\times 3$ for ROI Pooling 
since bounding boxes for people are 
\blue{typically tall}
and narrow.}

\begin{figure}[t]
	\centering
	\vspace{-2.5mm}
	\subfloat[]{\includegraphics[width=3.8cm,height=2.0cm]{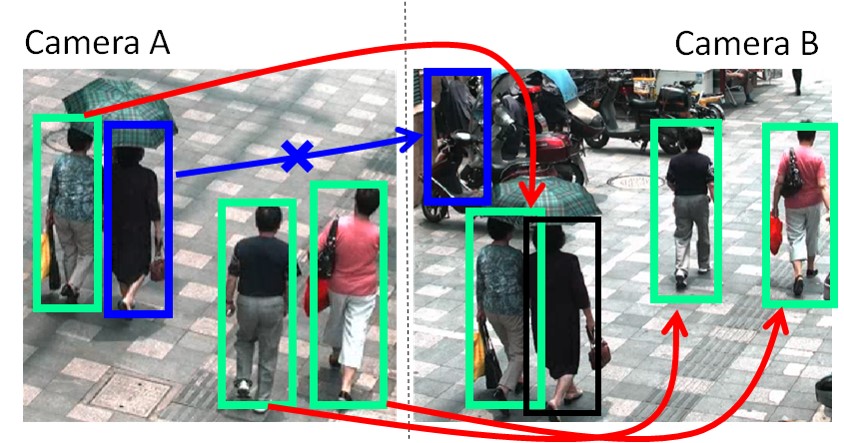}
		\label{fig:different mapping structures a}}
	{\vrule width0.6pt}
	\subfloat[]{\includegraphics[width=3.8cm,height=2.0cm]{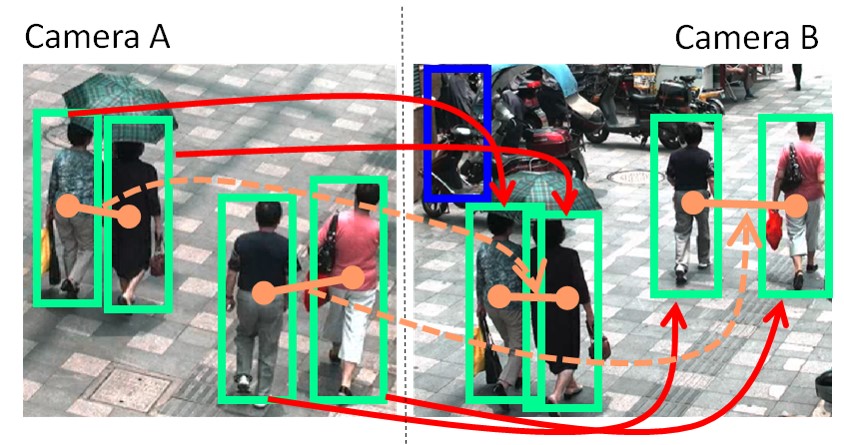}
		\label{fig:different mapping structures b}}\\
	\vspace{-4mm}
	\subfloat[]{\includegraphics[width=3.8cm,height=2.0cm]{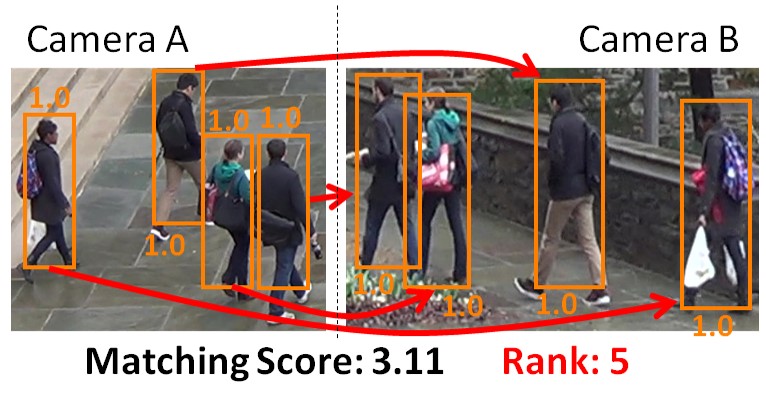}
		\label{fig:different mapping structures c}}
	{\vrule width0.6pt}
	\subfloat[]{\includegraphics[width=3.8cm,height=2.0cm]{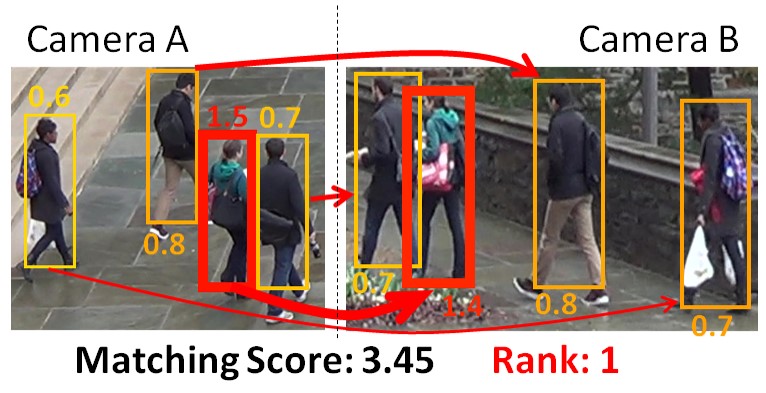}
		\label{fig:different mapping structures d}}\\
	\caption{Matching results by: (a) using only information of individuals; (b) using multi-grained information; (c) setting equal importance weights for all individuals/subgroups; (d) using our importance evaluation process to obtain importance weights. The red and blue links indicate correct and wrong matches, respectively. (Note: For a cleaner illustration, we only display the matching results between individuals.)}
	\label{fig:matching_results}
\end{figure}

\subsection{Ablation Studies}

\blue{We present our results alongside a number of ablation studies to provide a wide perspective of the performance of our proposed approach together with the impact of various components in the framework.}

\subsubsection{Results with features of different granularities}\label{sec:multi-grained feature}
In order to evaluate the effectiveness of our multi-grained group Re-ID framework, we compare eight methods \blue{of different granularities including some variations}: (1) Only using global features~\cite{mirror} of the entire group 
(\emph{Global}); (2) Only using features of individual people 
(\emph{Fine}); (3) Using features of individual people and two-people subgroup 
(\emph{Fine+Medium}); (4) Using features of individual, two-people, three-people subgroups 
(\emph{Fine+Medium+Coarse}); (5) Using our multi-grained framework, but \blue{assigning equal} importance weights for all people/people subgroups, i.e. set all to $1$ (\emph{Proposed-equal weights}); (6) Using our multi-grained framework, but \blue{omitting}
the spatial relation features in the multi-grained representation (cf. Sec.~\ref{section:feature extract}), \emph{Proposed-no spatial}); (7) Using our multi-grained framework, but using the ground truth pedestrian detection results (\emph{Proposed-GT}); (8) Using our multi-grained framework with automatic pedestrian detection method of~\cite{liulihang} to identify individual people in groups (\emph{Proposed-auto}).

Table~\ref{tab:cmcTable0} shows the CMC results of group Re-ID on the \emph{Road Group} dataset, measuring the correct match rates for different Re-ID ranks. 
The upper part lists the results based on hand-crafted features (cf. Sec.~\ref{sec:hand}) while the lower part lists the results based on deep convolutional features (cf. Sec. \ref{sec:cnn}).

Fig.~\ref{fig:matching_results} shows some group-wise matching results under different methods. We \blue{make the following observations:}

\begin{table}[t]
	\centering
	\caption{
	Ablation study results of representations of various granularities on the Road Group dataset.}
	\vspace{-3mm}
	\label{tab:cmcTable0}
	\footnotesize{
		\begin{tabular}{lccccc}
			\hline
			\rowcolor{mygray}
			\textbf{Rank (hand-crafted)} & 1& 5 & 10 & 15 & 20\\
			\hline
			Global & 15.8 & 31.6 & 43.0 & 48.6 & 54.8 \\
			Fine & 62.0 & 82.2 & 89.6 & 95.1 & 96.5 \\
			Fine+Medium & 66.7 & 87.2 & 93.3 & 96.0 & 96.8 \\
			Fine+Medium+Coarse & 71.1 & 89.4 & 94.1 & 97.0 & 97.3 \\
			\hline
			Proposed-equal weights& 55.8 & 78.0 & 88.1 & 92.1 & 93.6 \\
			Proposed-no spatial& {69.6} & {88.6} & {94.0} & {96.2} & {96.5} \\
			{\bf Proposed-auto}& {72.3}& {90.6}& {94.1}& {97.1}& {97.5}\\
			{\bf Proposed-GT}& {\bf 76.0} & {\bf 91.8} & {\bf 95.3} & {\bf 97.2} & {\bf 98.0} \\
			\hline
			\rowcolor{mygray}
			\textbf{Rank (deep convolution)} & 1& 5 & 10 & 15 & 20\\
			\hline
			Global & 32.1 & 67.9 & 77.8 & 84.0 & 86.4 \\
			Fine & 69.1 & 88.9 & 92.3 & 93.8 & 95.1 \\
			Fine+Medium & 69.1 & 90.1 & 95.1 & 96.3 & 96.3 \\
			Fine+Medium+Coarse & 72.8 & 93.8 & 95.1 & 96.3 & 96.8 \\
			\hline
			Proposed-equal weights& 72.4 & 90.1 & 92.6 & 96.3 & 97.5 \\
			Proposed-no spatial& {70.4} & {90.1} & {91.3} & {92.6} & {96.3} \\
			{\bf Proposed-auto}& {80.2}& {93.8}& {96.3}& {97.5}& {97.5}\\
			{\bf Proposed-GT}& {\bf 82.4} & {\bf 95.1} & {\bf 96.3} & {\bf 97.5} & {\bf 98.0} \\
			
			\hline
	\end{tabular}}
\end{table}

(1) The \emph{Global} method achieves poor results. This implies that simply using 
\blue{the entire group image}
cannot effectively handle the 
\blue{intricate variations that are present}
in group Re-ID. Comparatively, the \emph{Fine} method has obviously better performance by extracting and matching individual people to handle the challenges of group dynamics. However, its performance is still \blue{hindered} by the interference of pedestrian misdetections or \blue{mismatches} (cf. Fig.~\ref{fig:different mapping structures a}). These \blue{problems} are reduced \blue{more effectively} in the \emph{Fine+Medium} and \emph{Fine+Medium+Coarse} methods, \blue{both of} which, contain sub-group level information 
\blue{that captures underlying group dynamics.}
Our proposed framework (\emph{Proposed-GT} and \emph{Proposed-auto}), which includes all levels of granularity, 
can achieve the best performance.

(2) The \emph{Proposed-equal weights} method has obviously \blue{poorer} results than \blue{its counterpart}
with importance weights (\emph{Proposed-GT} and \emph{Proposed-auto}). This clearly indicates that: a) assigning importance weights to different \blue{individuals}/people subgroups is significant in guaranteeing group Re-ID performances; b) Our proposed importance evaluation scheme is effective in finding proper importance weights for all levels of granularity, such that reliable and discriminative individuals/people subgroups are highlighted, \blue{resulting in better matching results. } 
\redt{For instance, in Fig.~\ref{fig:different mapping structures c}-\ref{fig:different mapping structures d}, due to large layout change between group, the \emph{Proposed-equal weights} scheme \bluet{is unable to }
assign a high score on the pairs, while \emph{Proposed-auto} scheme allows
salient objects \bluet{to be given greater importance, hence resulting in a} higher matching score
}.

{(3) The \emph{Proposed-no spatial} method achieves relatively satisfactory results. This indicates that even when spatial relation features are not encoded, our approach can \blue{generally} still obtain reliable performances, \blue{propelled by} 
multi-grained information and importance weights. \redt{In the case of deep convolution features, \bluet{we observe a} relatively \bluet{larger} performance drop when spatial features are not used (about 10\% for rank-1), which indicates that the \bluet{choice of} spatial features extracted from our CNN \bluet{is crucial} 
and can boost the group Re-ID accuracy to a large extent.}}

(4) The \emph{Proposed-auto} method has \blue{almost} similar results to the \emph{Proposed-GT} method, \blue{only marginally lower in most cases.} \blue{The close performances of these two methods} 
indicate that our multi-grained group Re-ID framework has the ability to handle 
\blue{matching errors} caused by pedestrian misdetections. For example, in Fig.~\ref{fig:different mapping structures a}, the \blue{left} group in camera $A$ is \blue{incorrectly matched} 
with the blue 
\blue{rectangle}
in camera $B$ which detected a parked motorcycle. However, by integrating multi-grained information 
, we can successfully avoid this mismatch by considering subgroup correlation at higher-level granularities (cf. Fig.~\ref{fig:different mapping structures b}). \blue{Note that the iterative procedure refines the importance weights even at the fine granularity, subsequently resulting in a correct individual match.}

\red{(5) By comparing the results \blue{on a whole,}
we find a similar trend with respect to the combination of granularities, which \blue{affirms the good scalability and extensibility of} our group Re-ID framework \blue{to accommodate} 
different types of features. 
Further, we find when deep convolutional features are used, the performance of our framework is usually better than the hand-crafted counterpart on identical granularity settings.
}


\subsubsection{Results with different detection 
\blue{recalls}
}
\begin{figure}[]
	\centering
	\vspace{-4mm}
	\includegraphics[width=0.32\textwidth]{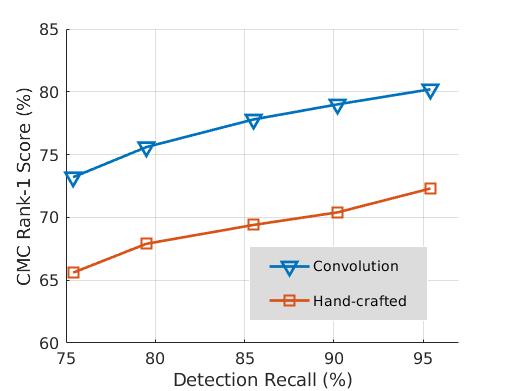}
	\caption{Rank-1 CMC scores via different detection recall rates using hand-crafted features (orange) and convolution features (blue) (Best viewed in color)}\label{fig:detection}
\end{figure}

\red{In Sec.~\ref{sec:multi-grained feature} we demonstrate that the matching accuracy is competitive 
\blue{when}
a high quality pedestrian detector \blue{is used}. Following this observation, we further investigate the effect of 
detection \blue{recalls} 
on our final matching accuracy. We conduct this experiment by \blue{altering} the output confidence threshold of our detector \cite{liulihang} to obtain detection results at different recall rates. \blue{We then perform group Re-ID based on the detected objects and their respective bounding boxes}  
In Fig.~\ref{fig:detection}, we compare the Rank-1 CMC scores for hand-crafted and deep convolutional features against detection recall rates.} 

\red{From Fig.~\ref{fig:detection}, we find the final matching results is relatively robust \blue{against} 
the quality of detectors that we adopt. \blue{This is evident as}
the drop in Rank-1 CMC score is less than $3\%$ for every corresponding $5\%$ decrease (\blue{approximately}) in recall rate. 
This observation further demonstrates that our multi-grained matching framework is robust to the detection quality and could still be helpful for group Re-ID problem even when the prior knowledge of individuals is incomplete.}
\begin{table*}[]
	\centering
	\caption{CMC results for group Re-ID on different datasets based on various feature combinations}
	\vspace{-3mm}
	\label{tab:cmcTable2}
	\footnotesize{
		\begin{tabular}{|l|c|c|c|c|c||c|c|c|c|c||c|c|c|c|c|}
			\hline
			\multirow{2}{*}{\diagbox{\textbf{Features}}{\textbf{Rank}}} & \multicolumn{5}{c||}{i-LIDS MCTS} & \multicolumn{5}{c||}{DukeMTMC Group} & \multicolumn{5}{c|}{Road Group}\\
			\cline{2-16}
			& 1& 5& 10& 15& 20 & 1& 5& 10& 15& 20 & 1& 5& 10& 15& 20\\
			\hline
			Hand-craft & 37.9 & 64.5 & 79.4 & 91.5 & 93.8 & 47.4 &68.1& 77.3 & 83.6 & 84.7
			& 72.3 & 90.6 & 94.1 & 97.1 & 97.5 
			\\
			Global-conv& 31.3 & 56.9 & 73.1 & 85.6 & 91.2 &  42.1& 67.8& 79.0 & 84.6 & 86.2
			& 75.3 & 93.8 & 96.3 & 96.3 & 97.5 
			\\
			Appearance-conv& 35.6 & 66.2 & 80.6 & 87.9 & 95.0 & 44.9 & 73.1 & 82.7	& 89.9	& 93.3
			& 76.8 & 92.3 & 95.1 & 97.5 & 97.5
			\\
			Full-conv& {\bf 38.8} & {\bf 65.7} & {\bf 82.5}	& {\bf 93.8} & {\bf 98.8} &  {\bf 48.4} & {\bf 75.2} & {\bf 89.9} & {\bf 93.3} & {\bf 94.4}
			& {\bf 80.2} & {\bf 93.8} & {\bf 96.3} & {\bf 97.5} & {\bf 97.5}
			\\
			\hline
	\end{tabular}}
	\vspace{-1em}
\end{table*}

\subsubsection{Results with different matching constraints}
In this study, we further evaluate the effectiveness of our multi-order matching process when 
\blue{matching constraints are varied. Five methods are compared:}
(1) Only single level matching, i.e., the first-order \blue{potential}
in \blue{Eq.~\ref{eq:weight11}} 
(\emph{Single)};
(2) \blue{Discard}
the 
inter-order potential \blue{term} i.e., setting all $\mathcal{\blue{P}}_{rl}(\mathcal{C})$ terms in Eq.~\ref{eq:objective} to $0$, (\emph{No inter}); 
(3) \blue{Discard}
the unmatched term (i.e., the second term in Eq.~\ref{equation:score}) when calculating matching scores (\emph{No dis}); 
(4) Use hyper-edge matching~\cite{RRWHM} 
, which integrates multi-grained information by constructing a \red{multi-order similarity function} (\emph{Hyp-E}); (5) Our proposed matching process (\emph{Proposed-auto}).

\red{Table~\ref{tab:cmcTable01} shows the CMC matching scores for group Re-ID task using multi-order matching with different matching constraints and \blue{criterion}. \blue{We report results for both} hand-crafted features and convolutional features on the \emph{Road Group} dataset.
From Table~\ref{tab:cmcTable01}, we can draw the following conclusions:}

(1) \blue{In comparison} with the \emph{Single} method, the \blue{higher-order potentials of the \emph{Proposed-auto} method clearly play an important role in} improving group Re-ID score \blue{by a great measure.} \blue{These potentials are essential to handle matching of multiple-person subgroups} to complement the use of multi-grained object representation 

(2) The \emph{Proposed-auto} method obtained better results than the \emph{No inter} method. This comparison indicates that the inter-order potential term is useful to properly 
\blue{capture correlations between different levels of granularity.}

(3) The \emph{Proposed-auto} method performed significantly better than the \emph{No dis} method. This demonstrates the importance of including the information of unmatched objects 
(cf. Eq.~\ref{equation:score}) \blue{in the matching process}. Results on both features show that this information has far more impact than that of the inter-order potential term (cf. method (2)).

(4) The \emph{Proposed-auto} method also has better matching accuracy 
than the \emph{Hyp-E} method. This demonstrates that our multi-order matching process can make better use of the multi-grained information in groups during matching.
\begin{table}[t]
	\centering
	\caption{Ablation study results of various matching order and criterion on the Road Group dataset}
	\vspace{-3mm}
	\label{tab:cmcTable01}
		\footnotesize{
			\begin{tabular}{lccccc}
				\hline
				\rowcolor{mygray}
				\textbf{Rank (hand-crafted)} & 1& 5 & 10 & 15 & 20\\
				\hline
				Single & 62.0 & 82.2 & 89.6 & 95.1 & 96.5 \\
				No inter & 70.1 & 88.8 & 94.1 & 96.3 & 97.5 \\
				No dis& 65.8 & 88.8 & 93.8 & 96.3 & 96.3 \\
				Hyp-E~\cite{RRWHM} & 55.1 & 77.8 & 87.6 & 88.9 & 95.1 \\
				{\bf Proposed-auto}& {\bf 72.3}& {\bf 90.6}& {\bf 94.1}& {\bf 97.1}& {\bf 97.5}\\
				\hline
				\rowcolor{mygray}
				\textbf{Rank (deep convolution)} & 1& 5 & 10 & 15 & 20\\
				\hline
				Single & 69.1 & 88.9 & 92.3 & 93.8 & 95.1 \\
				No inter & 74.1 & 92.6 & 96.3 & 97.5 & 97.5 \\
				No dis & 66.7 & 90.1 & 95.8 & 96.3 & 97.5 \\
				Hyp-E~\cite{RRWHM} & 62.9 & 77.8 & 88.9 & 95.1 & 96.3 \\
				{\bf Proposed-auto}& {\bf 80.2}& {\bf 93.8}& {\bf 96.3}& {\bf 97.5}& {\bf 97.5}\\
				
				\hline
		\end{tabular}}
\end{table}

\subsubsection{Results with different feature combination}
	
\red{We also investigate the effectiveness of using convolutional features from our multi-task CNN \blue{for different parts of the extracted features (see Fig.~\ref{fig:cnn}).} We evaluate the CMC score for all three datasets on the following combination of features: 
(1) Use hand-crafted features to describe both appearance and spatial relation for objects of all granulairity (\emph{Hand-crafted}). (2) Use convolutional features only to represent the global image and keep other features as hand-crafted (\emph{Global-conv}). (3) Represent appearance of global and local objects with convolutional features and keep spatial relation features as hand-crafted. (\emph{Appearnce-conv}). (4) Convolutional features \blue{for all parts} 
including the \blue{spatial relation} features.
(\emph{Full-conv}).}

\red{We report the matching accuracy with respect to different ranks in Table~\ref{tab:cmcTable2}. From these results, we can observe that:}

 \red{(1) \blue{Benefits are} 
 limited when only the global \blue{handcrafted} feature is replaced with one of deep convolutional. This \blue{shows that} 
 convolutional features are not \blue{sufficiently discriminative when applied on the entire group image.}}
 
 \red{(2) There \blue{is a leap in} improvement from \emph{Appreance-conv} to \emph{Full-conv} method, which indicates that using deep representations to encode \blue{spatial relations} between individuals is more 
 \blue{impactful}
 than opting for hand-crafted representations.}
 
 \red{(3) Overall, the improvement brought on by \emph{Full-conv} method over the \emph{Hand-crafted} method is \blue{more prominent} 
 on \emph{DukeMTMC Group} and \emph{Road Group} datasets than on \emph{i-LIDS MCTS}. This is \blue{indicative of the robustness of our proposed deep convolutional features
 in more \blue{crowded scenarios.} \pink{However, the handcrafted feature is still valuable since it can be applied on any scenarios without additional training process, especially when the data is limited and/or there are insufficient means to train a CNN feature extractor.}}
 }
\vspace{-3mm}
\subsection{Results 
on Single Person Re-ID}

\red{Although our approach is designed 
\blue{for} 
group Re-ID task, the intermediate result of fine-grained mapping $\mathcal{C}$ could be seen as a side product of our matching process. To further investigate how multi-order constraints and priors of group pairs effect the accuracy of individual matching, we conduct an extra experiment on single person Re-ID.}

\red{We compare two state-of-the-art single Re-ID methods, i.e. TriNet~\cite{Hermans2017Trihard} and AlignReID~\cite{Zhang2017AlignedReID},  
\blue{against} four 
variants of our matching scheme on \emph{Road Group} dataset: (1) Given a person from a probe image, we find the nearest person from \blue{among} all individuals in the gallery groups based on Euclidean distance in feature space (\emph{Single-match}). (2) Given a probe image, we first find its matched group in gallery, and for each individual in probe \blue{group} image, we find the nearest person \blue{from among the individuals in the matched gallery group} 
based on Euclidean distance in feature space (\emph{Intra-group}). (3) We first obtain the matched pairs $\mathcal{C}$ between groups by solving the group-wise multi-order matching problem, if the matched group is exactly the \bluet{ground} truth, we take $\mathcal{C}$ as matching results for individual objects in probe image, otherwise we regard all persons in probe groups as unmatched. (\emph{Proposed-auto}). 
We conduct these experiments on both hand-crafted and deep convolutional features.}

\begin{figure}[t]
	\centering
	\subfloat[]{\includegraphics[width=3.2cm,height=3.6cm]{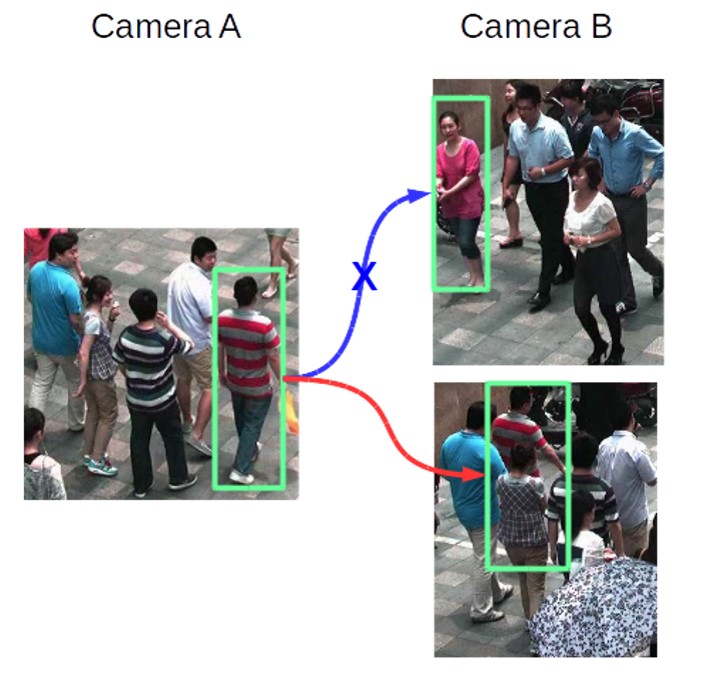}\label{fig:single-match1}}
	\hspace{0.35cm}
	{\vrule width0.6pt}
	\hspace{0.35cm}
	\subfloat[]{\includegraphics[width=3.2cm,height=3.6cm]{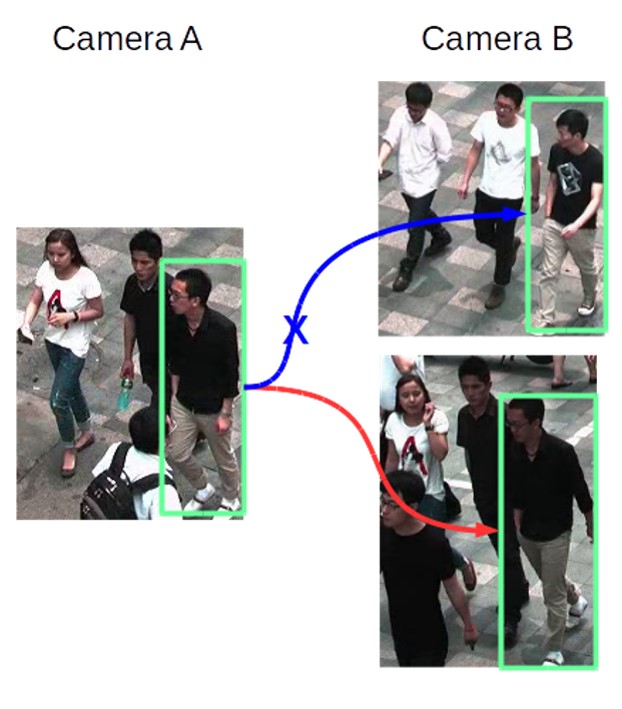}\label{fig:single-match2}}
	\caption{Examples of incorrect individual matching (blue arrows) under \emph{single-match} scheme and corresponding results under \emph{proposed} scheme (red arrows) (Best viewed in color).}\label{fig:single-match}
\end{figure}

\begin{figure}[t]
	\centering
	\vspace{-2.5mm}
	\subfloat[]{\includegraphics[width=3.8cm,height=2.0cm]{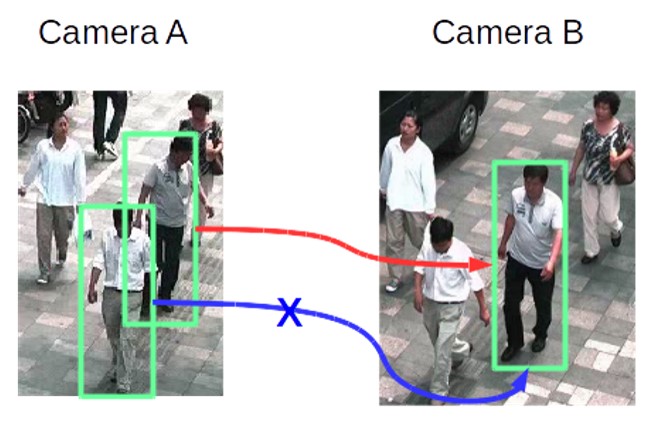}
		\label{fig:group mapping a}}
	{\vrule width0.6pt}
	\subfloat[]{\includegraphics[width=3.8cm,height=2.0cm]{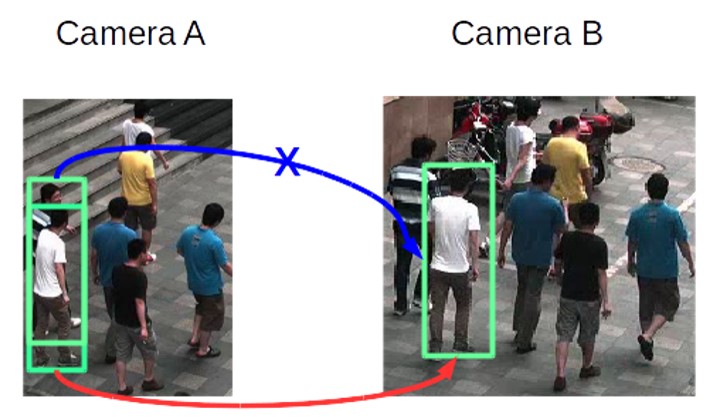}
		\label{fig:group mapping b}}\\
	\vspace{-4mm}
	\subfloat[]{\includegraphics[width=3.8cm,height=2.0cm]{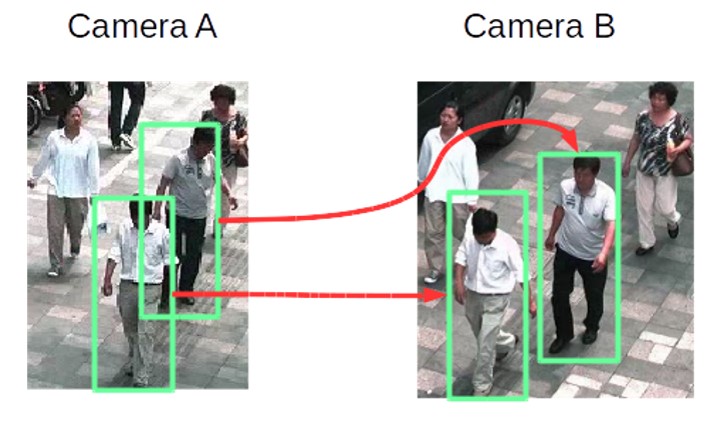}
		\label{fig:group mapping c}}
	{\vrule width0.6pt}
	\subfloat[]{\includegraphics[width=3.8cm,height=2.0cm]{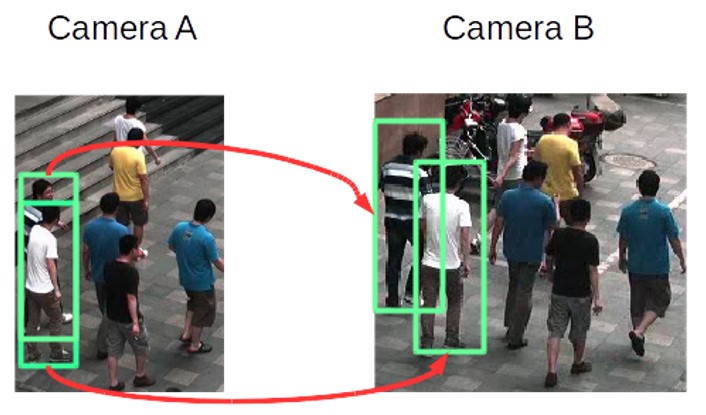}
		\label{fig:group mapping d}}\\
	\caption{Examples of individual matching results: (a) and (b) are matching results under \emph{intra-group} scheme, (c) and (d) are corresponding results under the \emph{proposed} scheme. The blue arrow indicates incorrect matches while red arrows denote the correct ones (best viewed in color).}
	\label{fig:group-single-re-id}
	\vspace{-2.5mm}
\end{figure}

\red{We report the 
Rank-1 CMC score \blue{for single person Re-ID} on the Road Group dataset in Table~\ref{tab:single-re-id} and further visualize some \blue{sample} results of different schemes in Fig.~\ref{fig:single-match} and Fig.~\ref{fig:group-single-re-id}. Table~\ref{tab:single-re-id} is split into two parts. The upper part lists results from methods without prior for groups, while the lower part lists results with group constraint. From these results, we observe the following:}

\begin{table}[t]
	\centering
	\caption{Rank-1 Results (R1) of different matching schemes for single person Re-ID on the Road Group Dataset}\label{tab:single-re-id}
	\footnotesize{
		\begin{tabular}{l|l|c}
			\hline
			 & Methods & \textbf{Rank-1 Score} \\
			\hline
			\multirow{4}{*}{Without Group} & TriNet~\cite{Hermans2017Trihard} & 37.2 \\
			& AlignReID~\cite{Zhang2017AlignedReID} & 32.8 \\
			& Single-match (hand-crafted) & 26.5  \\
			& Single-match (deep conv) & 21.0 \\
			\hline
			\hline
			\multirow{4}{*}{With Group} &
			Intra-group (hand-crafted) & 67.1 \\
			& Intra-group (deep conv) & 70.3 \\
			& {\bf Proposed-auto (hand-crafted)} &  71.4 \\
			& {\bf Proposed-auto (deep conv)} & {\bf 73.5} \\
			\hline
	\end{tabular}}
\end{table}


\red{(1) From Table~\ref{tab:single-re-id}, we observe that simple person-wise matching strategy without prior of groups performs rather poorly compared with other approaches. This indicates person Re-ID using only person-wise descriptors may be \blue{ill-suited} 
for such group scenarios since the \blue{search space} is likely too large \blue{with limited samples per person.} 
\blue{Without group priors, it is common to yield} matched individuals from other 
groups who are similar in appearance 
(illustrated by the blue arrows in Fig.~\ref{fig:single-match1}-\ref{fig:single-match2}).}

\red{(2) Although \emph{Intra-group} method performs better than person-wise matching, it is still \blue{worse} than the \emph{proposed-auto} scheme. Even by limiting the search space 
of the matched group, there still exists 
\blue{interferences} 
that results in \blue{incorrect individual matching}
For example, in Fig.~\ref{fig:group-single-re-id}, the matched results in Fig.~\ref{fig:group mapping a} 
\blue{show confusion}
with an individual \blue{with highly similar clothes}; in Fig.~\ref{fig:group mapping b}, there exists an occlusion \blue{in an overcrowded} area. Both of these factors result in failures in \emph{intra-group} method, while \emph{proposed-auto} scheme could handle these issues better since the constrain of multi-order potential requires the system to consider optimal matching not only between individuals but also between sub-groups of multiple person.}


\subsection{Comparison with State-of-the-art Methods}

    \begin{table*}[ht]
	\centering
	\caption{CMC results for group Re-ID on the three datasets}
	\vspace{-3mm}
	\label{tab:cmcTable1}
	\footnotesize{
		\begin{tabular}{|l|c|c|c|c|c||c|c|c|c|c||c|c|c|c|c|}
			\hline
			\multirow{2}{*}{\diagbox{\textbf{Method}}{\textbf{Rank}}} & \multicolumn{5}{c||}{i-LIDS MCTS} & \multicolumn{5}{c||}{DukeMTMC Group} & \multicolumn{5}{c|}{Road Group}\\
			\cline{2-16}
			& 1& 5& 10& 15& 20 & 1& 5& 10& 15& 20 & 1& 5& 10& 15& 20\\
			\hline
			Saliency~\cite{zhaorui2}& 26.1 & 48.5 & 67.5 & 80.3 & 89.9 & 13.9 & 33.3 & 51.5 & 59.8 & 66.3 & 48.6 & 73.6 & 82.2 & 86.2 & 90.1\\
			Mirror+KMFA~\cite{mirror}& 28.3 & 58.4 & 69.8 & 80.5 & 90.6 & 11.0 & 31.5 & 49.7 & 62.9 & 70.8 & 25.7 & 49.9 & 59.5 & 66.9 & 72.1\\
			CRRRO-BRO~\cite{zheng}& 23.3 & 54.0 & 69.8 & 76.7 & 82.7 & 9.9 & 26.1 & 40.2 & 54.2 & 64.9 & 17.8 & 34.6 & 48.1 & 57.5 & 62.2\\
			Covariance~\cite{cai}& 26.5 & 52.5 & 66.0 & 80.0 & 90.9 & 21.3 & 43.6 & 60.4 & 70.3 & 78.2 & 38.0 & 61.0 & 73.1 & 79.0 & 82.5 \\
			PREF~\cite{lisanti2017group}& 30.6  & 55.3 & 67.0 & 82.0 & 92.6 & 22.3 & 44.3 & 58.5 & 67.4 & 74.4 & 43.0 & 68.7 & 77.9 & 82.2 & 85.2 \\
			BSC+CM~\cite{icip16}& 32.0  & 59.1 & 72.3 & 82.4 & 93.1 & 23.1 & 44.3 & 56.4 & 64.3 & 70.4 & 58.6 & 80.6 & 87.4 & 90.4 & 92.1\\
			TriNet (local)~\cite{Hermans2017Trihard}& 25.0 & 53.2 &	65.6 & 78.2 & 84.4
			& 37.1 & 57.3 & 66.3 & 71.9 & 79.9
			& 67.8 & 87.7 & 88.9 & 93.8 & 96.3 
			\\
			AlignReID (local)~\cite{Zhang2017AlignedReID} & 28.1 & 56.3 & 68.8 & 75 & 87.5
			& 32.6 & 51.2 & 59.6 & 66.3 & 71.2
			& 69.8 & 87.4 & 94.1 & 94.1 & 96.3
			\\
			TriNet (global)~\cite{Hermans2017Trihard}& 33.6 & 55.0 &	69.4 & 77.5 & 86.9
			& 23.6 & 42.7 & 60.7 & 69.7 & 74.2
			& 34.6 & 65.4 & 82.7 & 82.4 & 90.2 
			\\
			AlignReID (global)~\cite{Zhang2017AlignedReID} & 34.4 & 62.5 & 75.0 & 84.3 & 93.7
			& 18.0 & 43.8 & 55.1 & 66.3 & 77.5
			& 39.5 & 55.6 & 70.4 & 77.8 & 85.9
			\\
			\hline
			{\bf Proposed-auto (hand)}& 37.9 & 64.5 &  79.4 & 91.5 &  93.8 &  47.4 &  68.1 &  77.3 &  83.6 &  87.4 &  72.3 &  90.6 &  94.1 &  97.1 &  97.5 \\
			{\bf Proposed-auto (conv)}&{\bf 38.8} & {\bf 65.7} & {\bf 82.5}	& {\bf 93.8} & {\bf 98.8} &  {\bf 48.4} & {\bf 75.2} & {\bf 89.9} & {\bf 93.3 } & {\bf 94.4}
			& {\bf 80.2} & {\bf 93.8} & {\bf 96.3} & {\bf 97.5} & {\bf 97.5}\\
			\hline
	\end{tabular}}
\end{table*}
Table~\ref{tab:cmcTable1} 
\blue{summarizes the group Re-ID performances on various 
datasets, comparing}
our proposed approach against state-of-the-art group Re-ID methods: \emph{CRRRO-BRO}~\cite{zheng}, \emph{Covariance}~\cite{cai}, \emph{PREF}~\cite{lisanti2017group}, \emph{BSC+CM}~\cite{icip16}. \blue{For clarity, we denote the features used in our method using the} \red{suffix \emph{hand} for
hand-crafted features and \emph{conv} 
for the \blue{convolutional features derived from our} multi-task deep CNN}.

For further \bluet{benchmarking} 
\red{we also include the results of state-of-the-art methods designed for single person Re-ID. \blue{Among them are} methods that utilize patch saliency (\emph{Saliency}~\cite{zhaorui2}) or a KMFA($R_{\chi^2}$) distance metric to calculate image-wise similarity (\emph{Mirror+KMFA}~\cite{mirror}). We also compare with two 
deep metric learning \blue{based methods}: TriNet\cite{Hermans2017Trihard}, a combination of CNN and triplet loss and AlignReID~\cite{Zhang2017AlignedReID}, a CNN-based method which simultaneously learns global and local distances between sample images. Since these two 
methods are \blue{originally} designed for person Re-ID, \red{we design two variants to} extend them for the group Re-ID scenario. One variant extracts features of individuals, under \bluet{their respective} deep frameworks \cite{Hermans2017Trihard,Zhang2017AlignedReID}, and \blue{proceeds to apply} Kuhn-Munkres' algorithm for bipartite matching between individuals in two groups. Finally, the similarity between 
two groups is computed as the inverse of the summation of feature distances between matched pairs. We denote this variant with a suffix \emph{local}, named after the \bluet{nature of this method.} The other variant directly takes the group image as the input of the algorithm in \cite{Hermans2017Trihard,Zhang2017AlignedReID} and calculates the group similarity according to Euclidean distance between output features. We denote this variant with a suffix \emph{global}, \bluet{since it considers the whole image.}} From Table~\ref{tab:cmcTable1}, we can observe that:

 (1) Our approach (handcrafted or deep convolutional features) has better results than the 
 \blue{other competing methods,}
 on all three evaluated datasets. This demonstrates the \blue{resounding consistency and} effectiveness of our approach \blue{in addressing the group Re-ID problem.} 
 
 (2) Group Re-ID methods that used global features (\emph{CRRRO-BRO}~\cite{zheng},  \emph{Covariance}~\cite{cai}, \emph{PREF}~\cite{lisanti2017group}) achieve less satisfactory results. This indicates that utilizing only global features is \blue{clearly inadequate at} handling the \blue{diverse range of} challenges in group Re-ID.
 
  (3) Although the \emph{BSC+CM} method obtained better results than that of global feature-based methods by introducing fine-grained objects (i.e. patches) to handle group dynamics, its performance is still \blue{evidently} lower than our approaches. This implies the usefulness of including \blue{information from multiple granularities}. 
  
  (4) Our approach is also obviously superior to  
  the conventional methods for single person Re-ID (\emph{Saliency}, \emph{Mirror+KMFA}). This indicates that the task of re-identifying each individual in a group-wise setting \blue{is a rather limited solution that is likely to fail in challenging} 
  group Re-ID scenarios. 
  
   \red{(5) Deep metric-learning methods perform poorer than our approach since they only resort to fine-grained representation (between person objects) while ignoring more complex patterns that occur at the medium and coarse sub-group levels. \blue{Interestingly,} this comparison also shows that these single Re-ID methods perform relatively better on groups with more individuals 
  (e.g. \emph{Road Group}).} 
   
    (6) The improvement of our approach is more obvious on datasets with larger group layouts and group member changes (\emph{DukeMTMC Group} \& \emph{Road Group}). This demonstrates that 
    our approach is \blue{capable of handling the dynamic changes that naturally occur in the group membership. \redt{On the other hand, the improvement on \emph{i-LDS MCTS} dataset is less obvious. This is the result of limited volume of people in this dataset, since for such scenarios, \bluet{representations based on multiple levels of granularity are} less
    discriminative 
    \bluet{than when applied to} crowded scenes; purely global descriptors appear to be sufficiently competitive for characterizing such small groups.}} 
    

\subsection{Computational Complexity}
\begin{table}[t]
\centering
\caption{Running time on the three datasets}\label{tab:time}
\footnotesize{
\begin{tabular}{|p{1.8cm}|c|*{5}{c|}c|}
\hline
\textbf{Datasets}& i-LIDS MCTS& DukeMTMC Group& Road Group\\
\hline
All image pairs& 1.1 min & 18.9 min & 11.5 min\\
Per image pair&  0.06 sec& 0.14 sec & 0.10 sec\\
\hline
\end{tabular}}
\end{table}

Table~\ref{tab:time} shows the running time of our group Re-ID approach on different datasets (excluding the time consumed for object detection and feature extraction). \pink{The running test is conducted on an 8-core i7-7700@3.60GHz CPU platform.}

We list two time complexity values: (1) the running time for the entire process (\emph{All image pairs in the dataset}), and (2) the average running time for computing the similarity of a single group image pair (\emph{Per image pair}). Table~\ref{tab:time} shows that our approach \pink{is acceptable in running time}.


\section{Conclusion\label{section:conclusion}}

This paper introduces a novel approach to address the seldom-studied problem of group re-identification. Our \blue{work contributes broadly in these aspects:} 
1) a multi-grained group Re-ID \blue{framework} which derives feature representations for multi-grained objects and iteratively evaluates their importance \blue{at different granularities} to handle 
group dynamics; 
2) a multi-order matching process which integrates multi-grained information to obtain more reliable group matching results;
\red{3) two independent pipelines (handcrafted and deep learning)}
\blue{which are capable of encoding appearance and spatial relations of multi-grained objects.}
Overall, our extensive experiments demonstrate the viability of our approaches. We also release our group Re-ID datasets involving realistic challenges to spur future works towards this direction.

\bibliographystyle{IEEEtran}
\bibliography{egbib}

\end{document}